%% file: example.tex
\documentclass[10pt,conference]{IEEEtran}

\usepackage{cite}

\usepackage{graphicx}
\usepackage[cmex10]{amsmath}
\usepackage{amsthm}

\usepackage{algorithmic}

\usepackage{array}

\ifCLASSOPTIONcompsoc
  \usepackage[caption=false,font=normalsize,labelfont=sf,textfont=sf]{subfig}
\else
  \usepackage[caption=false,font=footnotesize]{subfig}
\fi
\usepackage{url}

\usepackage{enumerate}

\usepackage[caption=false]{subfig}

\usepackage{amssymb}
\usepackage{url}
\usepackage[acronym]{glossaries}
\usepackage{multirow}
\usepackage{makecell}
\usepackage{placeins}
\usepackage{xcolor}
\definecolor{yellowContrast}{RGB}{200,140,0}
\usepackage{supertabular,booktabs}
\usepackage{tabularx}
\usepackage{balance}
\usepackage{longtable}
\usepackage{enumitem}
\newcolumntype{Y}{>{\centering\arraybackslash}X}

\newacronym{cnn}{CNN}{Convolutional Neural Network}
\newacronym{knn}{kNN}{K-Nearest Neighbors}
\newacronym{sod}{SOD}{Salient Object Detection}
\newacronym{gan}{GAN}{Generative Adversarial Network}
\newacronym{lbp}{LBP}{Local Binary Patterns}
\newacronym{fcn}{FCN}{Fully Convolutional Network}
\newacronym{maskrcnn}{Mask-RCNN}{Mask Regional Convolutional Neural Network}
\newacronym{resnet}{ResNet}{Residual Network}
\newacronym{res2net}{Res2Net}{Res2Net}
\newacronym{mae}{MAE}{Mean Absolute Error}

\usepackage{tikz}

\newcommand\copyrighttext{%
  \footnotesize \textcopyright 2020 IEEE. Personal use of this material is permitted.
  Permission from IEEE must be obtained for all other uses, in any current or future
  media, including reprinting/republishing this material for advertising or promotional
  purposes, creating new collective works, for resale or redistribution to servers or
  lists, or reuse of any copyrighted component of this work in other works.
  }
\newcommand\copyrightnotice{%
\begin{tikzpicture}[remember picture,overlay]
\node[anchor=south,yshift=10pt] at (current page.south) {\fbox{\parbox{\dimexpr\textwidth-\fboxsep-\fboxrule\relax}{\copyrighttext}}};
\end{tikzpicture}%
}

\begin{document}
%
\title{IDA: Improved Data Augmentation Applied to Salient Object Detection}

\newif\iffinal
\finaltrue
\newcommand{\cmtid}{15}


\iffinal


\author{\IEEEauthorblockN{Daniel V. Ruiz and Bruno A. Krinski and Eduardo Todt}
\IEEEauthorblockA{Department of Informatics, Federal University of Parana, Brazil\\
Email: \{dvruiz, bakrinski, todt\}@inf.ufpr.br}}


%

\else
  \author{Sibgrapi paper ID: \cmtid \\ }
\fi

\maketitle
\copyrightnotice

\begin{abstract}
In this paper, we present an Improved Data Augmentation~(IDA) technique focused on Salient Object Detection~(SOD).  Standard data augmentation techniques proposed in the literature, such as image cropping, rotation, flipping, and resizing, only generate variations of the existing examples, providing a limited generalization. Our method combines image inpainting, affine transformations, and the linear combination of different generated background images with salient objects extracted from labeled data.  Our proposed technique enables more precise control of the object's position and size while preserving background information. The background choice is based on an inter-image optimization, while object size follows a uniform random distribution within a specified interval, and the object position is intra-image optimal. We show that our method improves the segmentation quality when used for training state-of-the-art neural networks on several famous datasets of the SOD field. Combining our method with others surpasses traditional techniques such as horizontal-flip in 0.52\% for F-measure and 1.19\% for Precision. We also provide an evaluation in 7 different SOD datasets, with 9 distinct evaluation metrics and an average ranking of the evaluated methods.
\end{abstract}


\IEEEpeerreviewmaketitle

\input{main}

\bibliographystyle{IEEEtran}
\bibliography{example}
%
%


\end{document}

%% file: main.tex
\section{Introduction}\label{sec:intro}

A visual scene is a complex structure composed of many different regions and objects, with a large variety of color, size, and texture. The human visual system has the natural ability to filter such complex structures and focus on the most attended regions, also called salient regions, making it faster for our brain to analyze and understand all regions and objects located in the scene. Aiming to replicate this ability in machines the Salient Object Detection~(SOD) research field focuses on studying Computer Vision techniques to find the most salient objects in images~\cite{ssvatunc}.

In recent decades, the SOD literature presented an impressive growth in the number of novel and promising approaches. Recent works, which are based on Deep Learning techniques, have shown remarkable results in the field~\cite{wang2019sodsurvey, krinski2019}. Due to its high precision and generalization abilities, Deep Learning-based methods can find the salient regions of images with higher reliability. However, such methods weak point is the amount of data required to train them. For that reason, popular datasets such as ImageNet~\cite{imagenet} and PLACES~\cite{zhou2018places2} are composed of millions of images.

Moreover, besides obtaining a significant number of images to compose a dataset in a supervised learning approach, the images have to be labeled. The labeling of images is specialized and laborious work and can be even more expensive in segmentation tasks like SOD, which requires a pixel-wise segmentation mask for each image in the dataset.

In general, two approaches are usually applied to bypass such a necessity of data: transfer learning and data augmentation. The former makes use of pre-trained models in more massive datasets, like ImageNet. In this strategy, the objective is to use generic features already learned by the model and then fine-tuning it to the proposed problem with small datasets~\cite{htafddn}. The latter makes use of different augmentation techniques like horizontal/vertical flip, image cropping, rotation, and others, to synthetically generate new training samples based on the existing ones~\cite{da}. While others deal with domain adaptation to expand the data available in novelty ways as in
~\cite{ruiz2020giraffes}. In this work, we address the data problem by proposing a technique called Improved Data Augmentation~(IDA).

The SOD problem is addressed in the literature as a cross-testing problem~\cite{borji:survey}, with the training being performed in one dataset and the testing being performed in other datasets, utterly different from the training dataset. When analyzing the SOD datasets proposed in the literature, we noticed that the training dataset is not generic enough. In general, the training dataset contains certain biases such as large salient objects positioned in the center of the image, with size and position distribution very different from the testing datasets. Thus, our primary goal is to generate new salient examples in compliance with a more widespread distribution that can encompass the testing datasets' distribution to train more precise models and prevent overfitting to the original training dataset.

In this paper, we propose an improved approach to data augmentation in the context of SOD. Our method uses a linear combination of two different images. The resulting image contains in the foreground: a salient object segmented from its original background subsequently affinely transformed, and a full background generated created using image inpainting to erase its labeled objects. Our proposed method can be summarized in five steps: background generation, feature extraction, distance optimization, affine transformation, and linear combination.
The implementation of the method is publicly available~\footnote{\url{https://github.com/VRI-UFPR/IDA}}.

To demonstrate the effectiveness of this technique, we present quantitative and qualitative results obtained by augmenting the public available dataset DUTS~\cite{wang2017duts} to train the state-of-the-art deep neural network PoolNet~\cite{liu2019poolnet} with the \gls{res2net} backbone~\cite{gao2019res2net} for \gls{sod}. 

\section{Related Work}\label{sec:related}

The majority of data augmentations techniques proposed in the \gls*{sod} literature are restricted to affine transformations, random crop, and random flip. Perazzi \emph{et al.}~\cite{Perazzi2017CVPR} proposed affine transformations to generate new images by scaling and translating the training images. Guo \emph{et al.}~\cite{GuoSymmetry2018} and Liu \emph{et al.}~\cite{liu2019poolnet} used random flip as data augmentation. Wei \emph{et al.}~\cite{wei2019f3net}, and Wang \emph{et al.} \cite{wang2019progressive} utilized image resize, random crop and random horizontal flip.

To the best of our knowledge, Ruiz \textit{et al.}~\cite{ruiz2019anda} proposed the first method incorporating an inpainting technique for data augmentation of salient objects images. However, this method, named ANDA, does not guarantee intra-image optimal salience, since object placement simply follows a random uniform distribution.

Chen et al.~\cite{chen2020gridmask} proposed a data augmentation method named GridMask based on structured information removal. The proposed method generates a set of binary masks with a sequence of square blocks, with pixel value equal to 0, uniformly distributed over the masks in a grid structure. The squares have the same size and same distance between them. When a mask is applied over the image, the image's regions corresponding to a square region in the GridMask are erased, generating a black region in the image without any information. Chen et al.~\cite{chen2020gridmask} demonstrated that the proposed method could be successfully applied to different deep learning problems like image classification, object detection, and semantic segmentation. Also, they showed that GridMask outperforms popular erase-based data augmentation techniques like random erasing~\cite{zhong2020random}, CutOut~\cite{devries2017cutout}, and hide-and-seek (HaS)~\cite{singh-arxiv2018}.

Recent advances in backbone \glspl{cnn} has improved multi-scale representation, leading to consistent performance gains on a wide range of applications. However, most existing methods represent the multi-scale features in a layer-wise manner. Recently, Gao et al.~\cite{gao2019res2net} proposed a novel building block for \glspl{cnn}, a \gls{resnet}, namely Res2Net, that can represent multi-scale features at a granular level and increases the range of receptive fields for every network layer. This was achieved by constructing hierarchical residual-like connections within one single residual block. In their work~\cite{gao2019res2net}, Gao et al. demonstrated that this backbone provides consistent performance gains in multiple computer vision tasks, such as object detection, semantic segmentation, and salient object detection.

To evaluate the Res2Net in SOD, Gao et al.~\cite{gao2019res2net} replaced the \gls{resnet}-50 backbone in the state-of-the-art Neural Network PoolNet~\cite{liu2019poolnet} and kept all the other configurations unchanged. The authors then trained a baseline model with the \gls{resnet}-50 and a Res2Net model using the MSRA-B dataset~\cite{liu2007msrab}. The Res2Net model achieved better results when compared with the \gls{resnet}-50 in four famous \gls{sod} datasets: ECSSD, PASCAL-S, HKU-IS, and DUT-OMRON.

Proposed by Liu et al.~\cite{liu2019poolnet}, the PoolNet is based on a Feature Pyramid Network (FPN)~\cite{lin2017fpn}. Liu et al.~\cite{liu2019poolnet} also showed that their PoolNet architecture achieves better results with the \gls{resnet}-50 backbone than with the VGG-16. Thus, discouraging future works using the PoolNet architecture with a VGG-16 backbone.  Following ~\cite{gao2019res2net,ruiz2019anda}, we focus on the PoolNet model without joint training with edge detection.

\section{Proposed Work}\label{sec:work}

\begin{figure*}[!htb]
\centering
\includegraphics[width=0.9\linewidth]{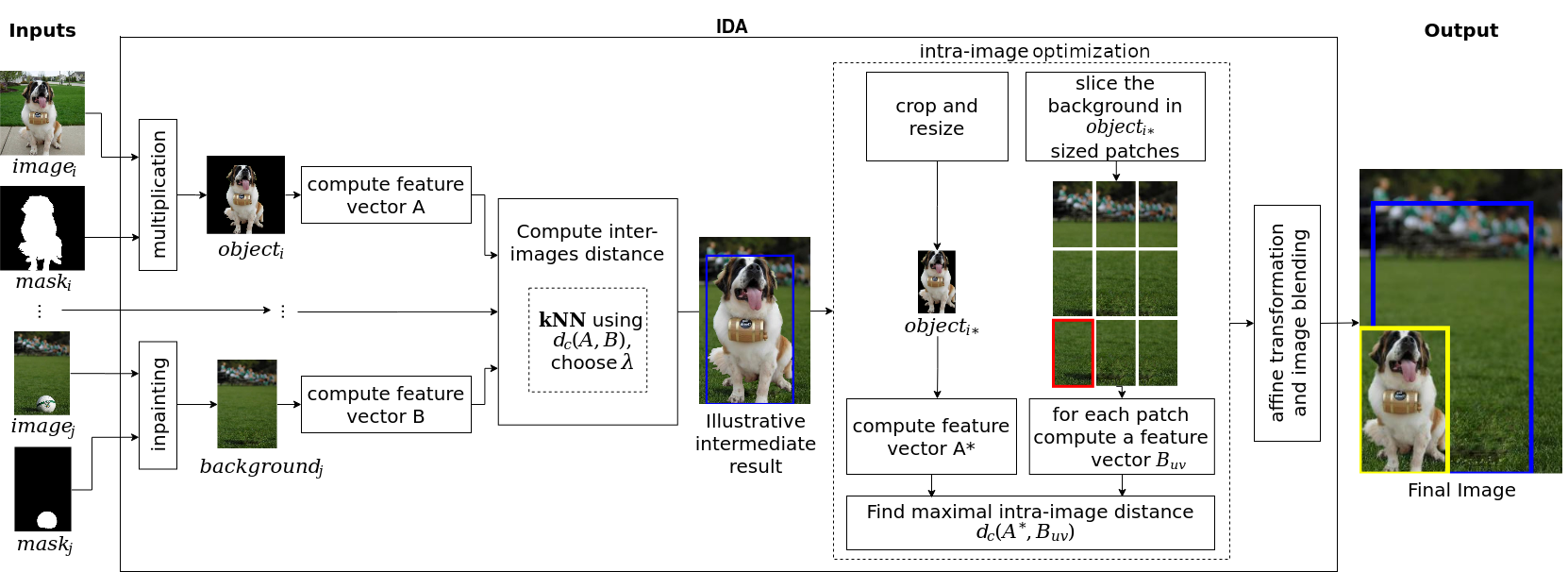}
\caption{Flowchart of the IDA technique. The first step is the computation of a background image for all the images of the inputted dataset; an object image is also computed; those images are used to produce feature vectors that can be compared using kNN with cosine similarity; $\lambda$ stands for the criteria chosen, further details on Section~\ref{sec:linearObjBg}. Highlighted in \textbf{\textcolor{blue}{blue}} is the intermediate bounding box of the object before the affine transformation. An intra-image optimization determines the translation of the object in the new background. Highlighted in \textbf{\textcolor{red}{red}} is the patch that produces the maximal distance intra-image in this example. Finally, for each pair of $object_i$ and $background_j$, a final image is created using a linear combination of both.  Highlighted in \textbf{\textcolor{yellowContrast}{yellow}} is the final bounding box of the object.}
\label{fig:flowchart}       
\end{figure*}

Our proposed method comprises five steps: background generation, feature extraction, distance optimization,  affine transformation, and linear combination. Fig.~\ref{fig:flowchart} presents the complete pipeline of our method. 

In the first step, we use an inpainting method to remove the salient object from an image altogether, maintaining only the background of the image (details presented in  Section~\ref{sec:backgroundImage}). We then extract color and texture features from each new background image and each of the dataset's initial salient objects. In the third step, we compute the distance between the feature vectors to perform an inter-image optimization using the \gls{knn} algorithm and cosine similarity to match an object and a background (details presented in  Section ~\ref{sec:linearObjBg}). Later on, an intra-image optimization is performed to match a patch of the background and the object.

In the last step, we apply affine transformations in the salient object. The transformations are based on the size and position distribution of the original dataset objects. Then, a linear combination is performed to blend the transformed salient object in the new background, thus generating a new image (details presented in  Section ~\ref{sec:size_position}).

\begin{figure}[htb]

\centering
\includegraphics[width=\linewidth]{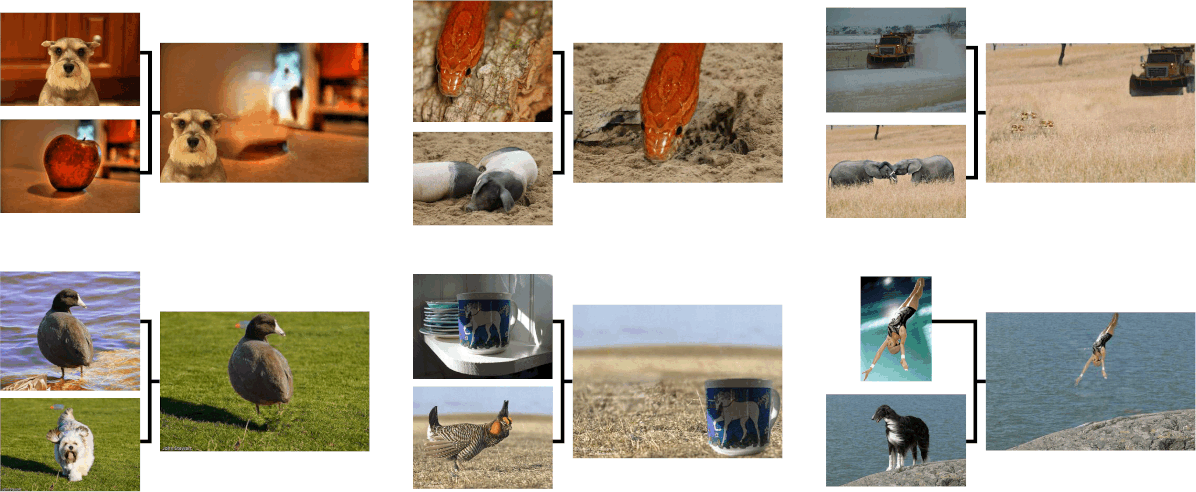}

\caption{Six examples of generated images by IDA. In each bracket, the new image (on the right) was generated with the object of the top-left image and the background of the bottom-left image.}
\label{fig:examples}       
\end{figure}

To showcase what the resulting synthetic images look like, we present in Figure~\ref{fig:examples} six different examples of the images produced by the IDA method, with input images from the DUTS-TRAINING set. The selection of background images was based on the inter-image distance computation step described in Section~\ref{sec:linearObjBg}.

\subsection{Background Image Generation}
\label{sec:backgroundImage}

The first step of the method is to generate background images, without any salient object, for each image in the training dataset. The idea is that each labeled image can produce both a foreground and a background sample. The foreground image retains only the object of interest, and the background image retains all other information except the data of the object. This introduced void is filled with a technique called image inpainting, producing a complete image. 

\begin{figure}[!htb]
\centering

\subfloat[Image]{\includegraphics[width=0.245\linewidth]{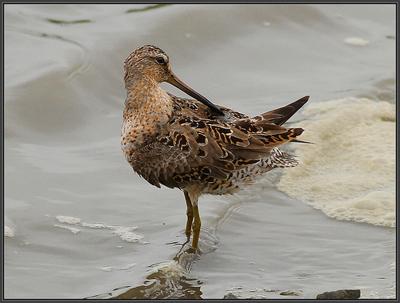}}
\subfloat[Mask]{\includegraphics[width=0.245\linewidth]{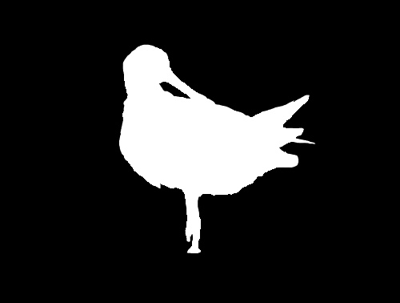}}
\subfloat[DeepFill v2]{\includegraphics[width=0.245\linewidth]{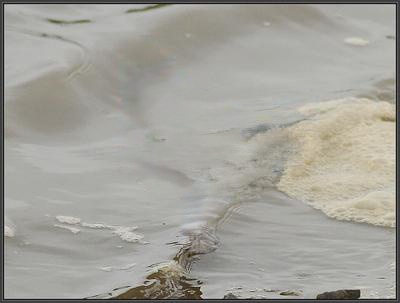}\label{subfig:newinpaint_bird}}
\subfloat[Adapted Pconv]{\includegraphics[width=0.245\linewidth]{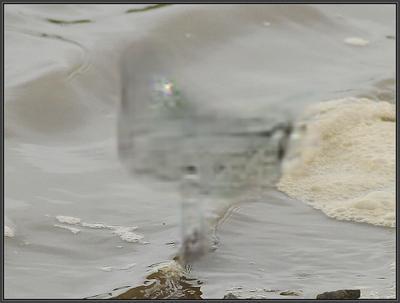}\label{subfig:oldinpaint_bird}}

\subfloat[Image]{\includegraphics[width=0.245\linewidth]{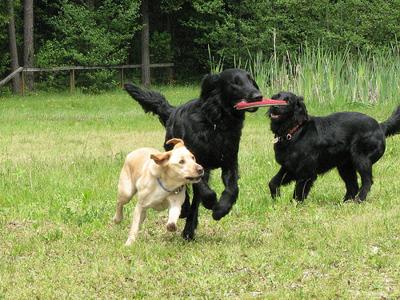}}
\subfloat[Mask]{\includegraphics[width=0.245\linewidth]{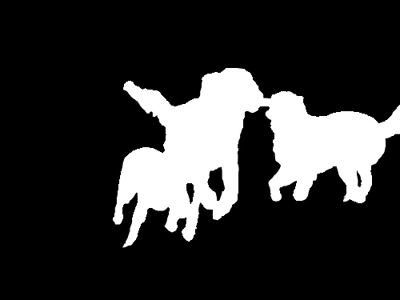}}
\subfloat[DeepFill v2]{\includegraphics[width=0.245\linewidth]{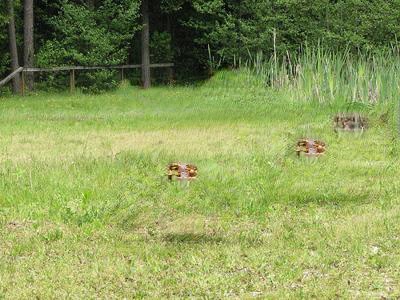}\label{subfig:newinpaint_dogs}}
\subfloat[Adapted Pconv]{\includegraphics[width=0.245\linewidth]{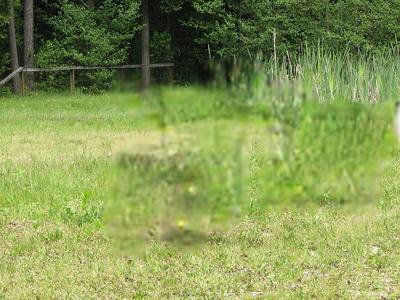}\label{subfig:oldinpaint_dogs}}

\caption{Qualitative difference between inpainted performed by the previous method Adapted Pconv~\cite{ruiz2019anda}~(Subfigures~\ref{subfig:oldinpaint_bird},\ref{subfig:oldinpaint_dogs}) and the new one DeepFill v2~\cite{yu2018CVPR,yu2019ICCV}~(Subfigures~\ref{subfig:newinpaint_bird},\ref{subfig:newinpaint_dogs}). Note how the DeepFill v2 method produces a more detailed erasure while sometimes producing an artifact~(Subfigure~\ref{subfig:newinpaint_dogs}) on the image.}
\label{fig:inpaint}       
\end{figure}

Image inpainting is the process of restoring missing pixels of digital images plausibly, reconstructing a region based on the surrounding information. Recently, research in image inpainting has received considerable attention in different areas, and it can be used in many different applications, such as restoration of old and damaged documents, removal of unwanted objects, and retouching~\cite{qureshi2017inpaintsurvey}. Previous works~\cite{ruiz2019anda} used an UNet-like architecture~\cite{unet2015} named PConv~\cite{nvidiaInpainting} to construct background images without salient regions within. We use a neural network called DeepFill v2~\cite{yu2018CVPR,yu2019ICCV}, officially available on\footnote{\url{https://github.com/JiahuiYu/generative_inpainting}} and pre-trained with the Places2 dataset~\cite{zhou2018places2}. The DeepFill v2 achieved better inpainting results than the adapted version of PConv, as presented in Fig.~\ref{fig:inpaint}.

\subsection{Feature Extraction and Inter-image Optimization}
\label{sec:linearObjBg}
After generating images without any salient object, our method performs the selection of a new salient object to be inserted in the generated background image. However, when a salient object is inserted in a new random background, the salient object may no longer be salient. Our method chooses a background that retains some of the object's salience to overcome such a problem. To find such a background, we compute a feature vector of 256 positions composed of four histograms for both the salient object and the new background. To represent the color, we employ an HSV color space histogram divided into 64 bins for Hue, 64 for Saturation, 64 for Value, or Brightness. To represent texture, we employ a \gls*{lbp}~\cite{Ojala2002lbp} histogram with 64 bins, with the following parameters: the number of circularly symmetric neighbor points set to 24, the circle radius set to 3, and the uniform method. 

Then, the \gls*{knn} algorithm is applied to measure each salient object's distance to each inpainted background. We use ${k=10,553}$, in which $10,553$ is the number of images in the DUTS-TRAINING dataset. In the \gls*{knn}, we use the cosine similarity defined by Equation~\ref{eq:cosine}. The similarity value is obtained using the feature vector A originated from the salient object without its original background, and the feature vector B originated from the new background image. In this way, $d_c(A,B)$ shows the similarity between the object and the background. 

\begin{equation}
    d_c(A,B)=\frac{\sum_{j=1}^{256}A_jB_j}{\sqrt{\sum^{256}_{j=1}A_{j}^{2}}\sqrt{\sum^{256}_{j=1}B_{j}^{2}}}
\label{eq:cosine}
\end{equation}

\begin{figure}[!htb]
\centering

\subfloat[Mask]{\includegraphics[width=0.32\linewidth]{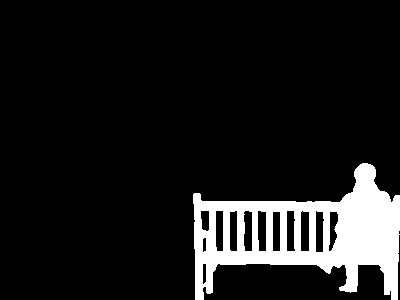}\label{subfig:mmask_fixed}}
\hfill
\subfloat[$\lambda=b_{\left \lfloor{ \frac{k}{2} }\right \rfloor}$]{\includegraphics[width=0.32\linewidth]{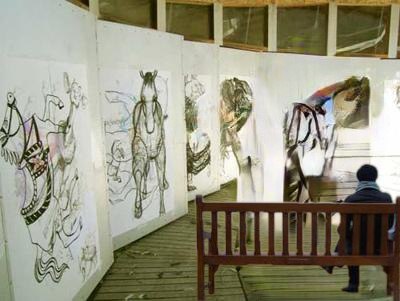}\label{subfig:fixed}}
\hfill
\subfloat[$\lambda=b_{c}$]{\includegraphics[width=0.32\linewidth]{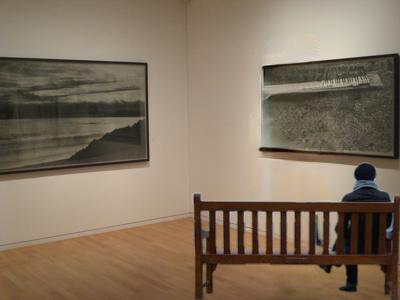}\label{subfig:closest}}
\caption{Qualitative impact of the background choosing. $\lambda$ stands for the criteria chosen, $b$ stands for the neighbor chosen. Note how the introduced object is more salient in Subfigure~\ref{subfig:closest} than in Subfigure~\ref{subfig:fixed}.}
\label{fig:backgroundcriteria}       
\end{figure}

Given an object $o$, let $\mu_o$ be the mean similarity value of the \glspl*{knn} of $o$, and $\sigma_o$ the similarity standard deviation of the \glspl*{knn} of $o$. Instead of using the $\left \lfloor{ \frac{k}{2} }\right \rfloor$th neighbor for all the images as in~\cite{ruiz2019anda}, a dynamic choice for each image based on the closest neighbor to $\mu_o+\sigma_o$ can better preserve the salience on the new samples. Fig.~\ref{fig:backgroundcriteria} illustrates the difference when using the $\left \lfloor{ \frac{k}{2} }\right \rfloor$ and the $c=closest(\mu_o+\sigma_o)$ criteria.

\subsection{Size distribution and intra-image position optimization}
\label{sec:size_position}

\begin{figure}
\centering
\captionsetup[subfigure]{labelformat=empty,position=top}

\subfloat[Position Distribution]{\includegraphics[width=0.35\linewidth]{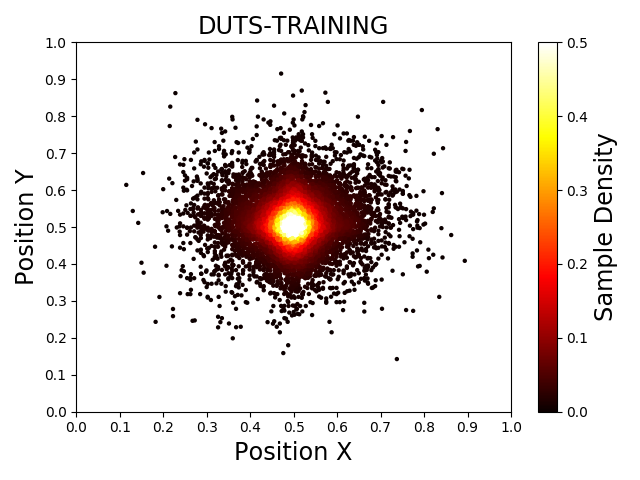}}
\hspace{0.5cm}
\subfloat[Size Distribution]{\includegraphics[width=0.35\linewidth]{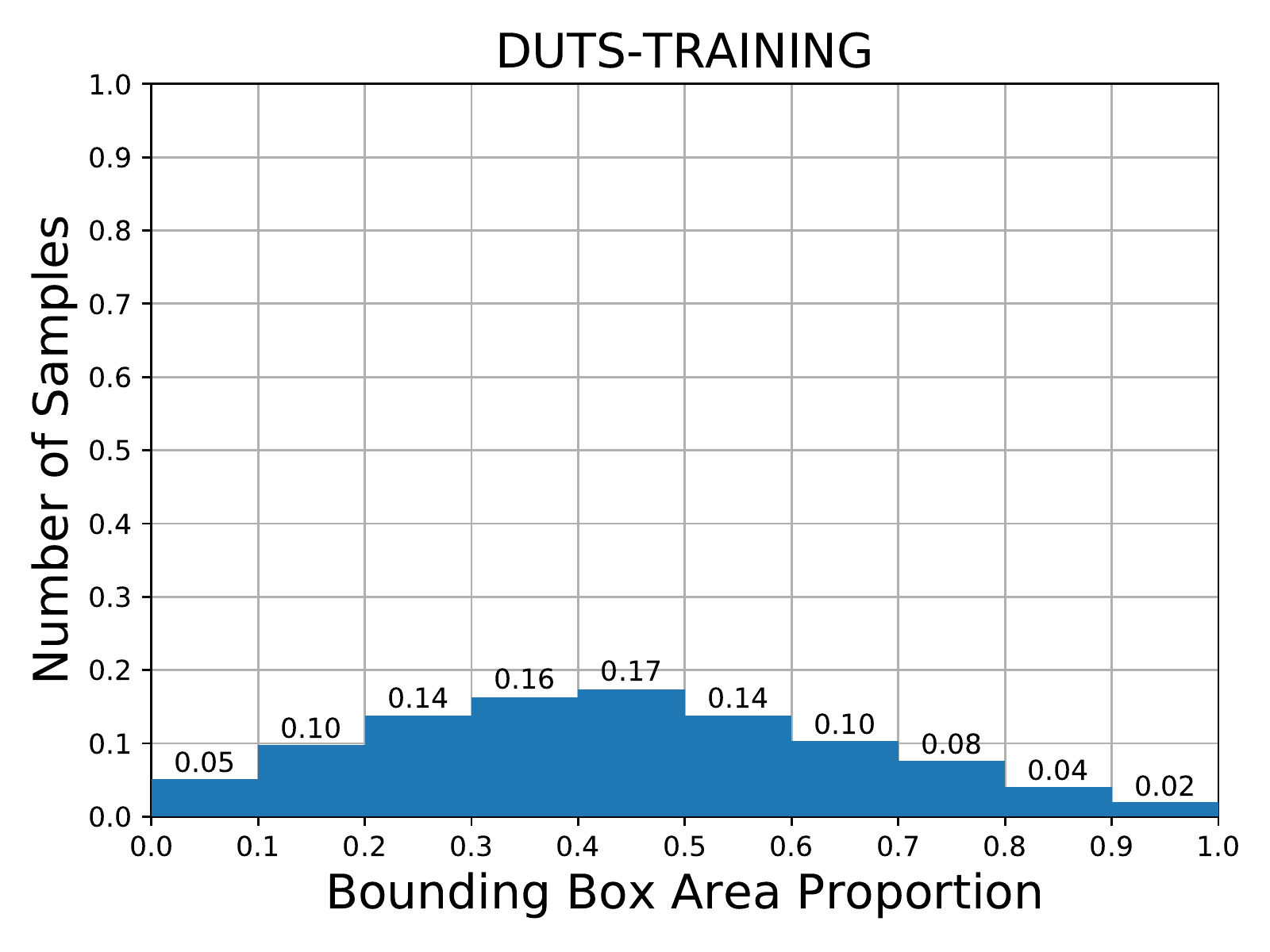}}

\subfloat{\includegraphics[width=0.35\linewidth]{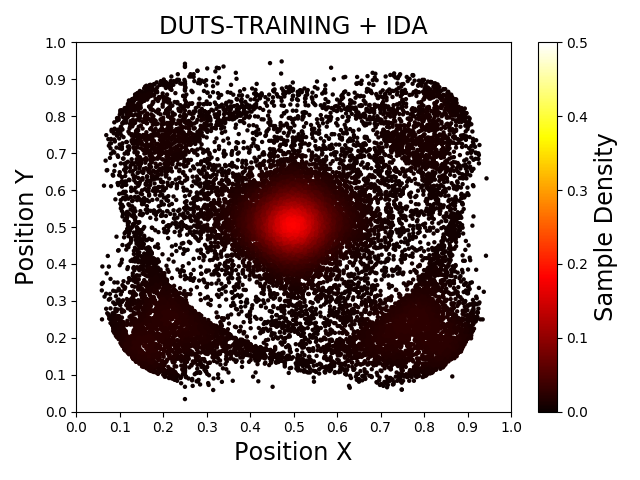}}
\hspace{0.5cm}
\subfloat{\includegraphics[width=0.35\linewidth]{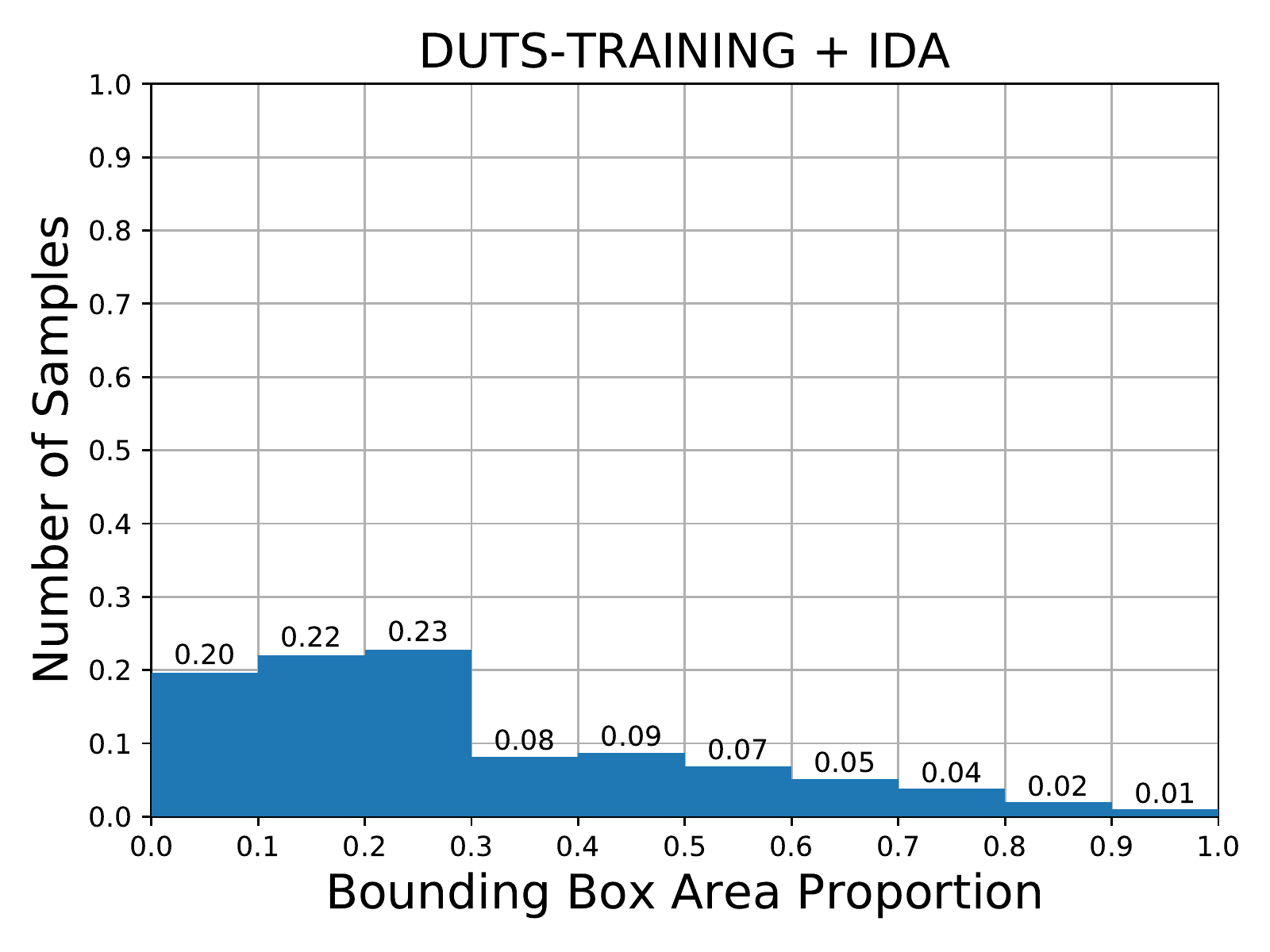}}

\caption{Position and size distribution per dataset for the training: DUTS-TRAINING, DUTS-TRAINING+IDA. The position distribution is demonstrated in a scatter plot of the normalized bounding box center coordinates. Additionally, a heat colormap represents the sample position density. For size, the bounding box area divided by image area is displayed in a 10-bin histogram.}
\label{fig:dist1}       
\end{figure}

\begin{figure}
\centering
\captionsetup[subfigure]{labelformat=empty,position=top}

\subfloat[Position Distribution]{\includegraphics[width=0.35\linewidth]{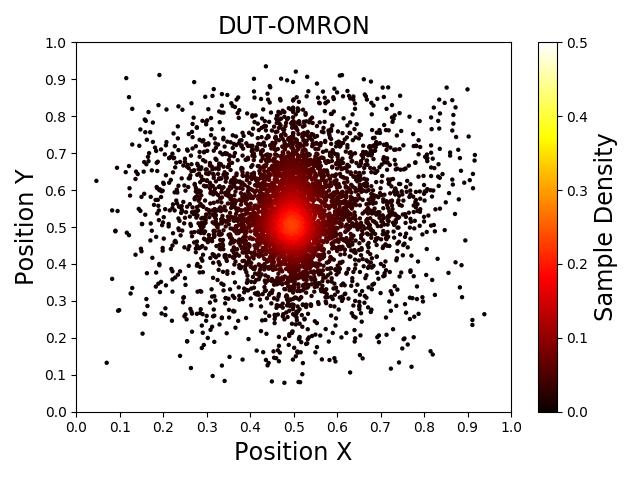}}
\hspace{0.5cm}
\subfloat[Size Distribution]{\includegraphics[width=0.35\linewidth]{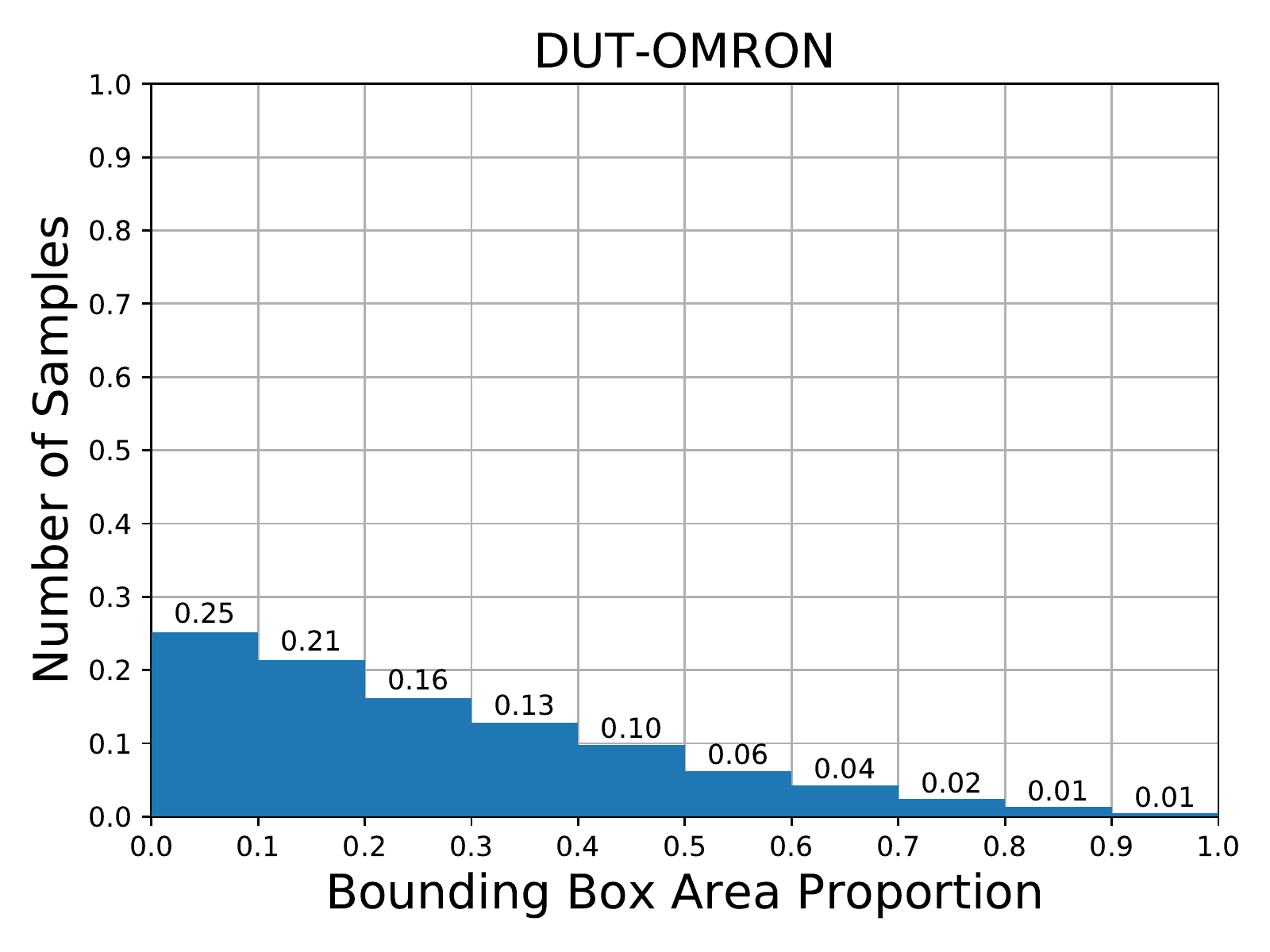}}

\subfloat{\includegraphics[width=0.35\linewidth]{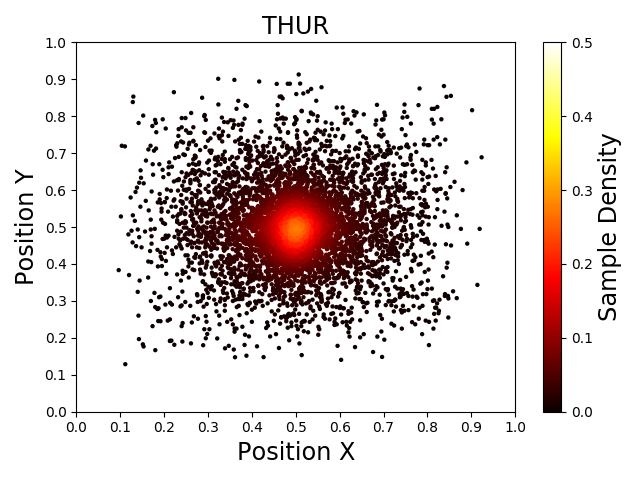}}
\hspace{0.5cm}
\subfloat{\includegraphics[width=0.35\linewidth]{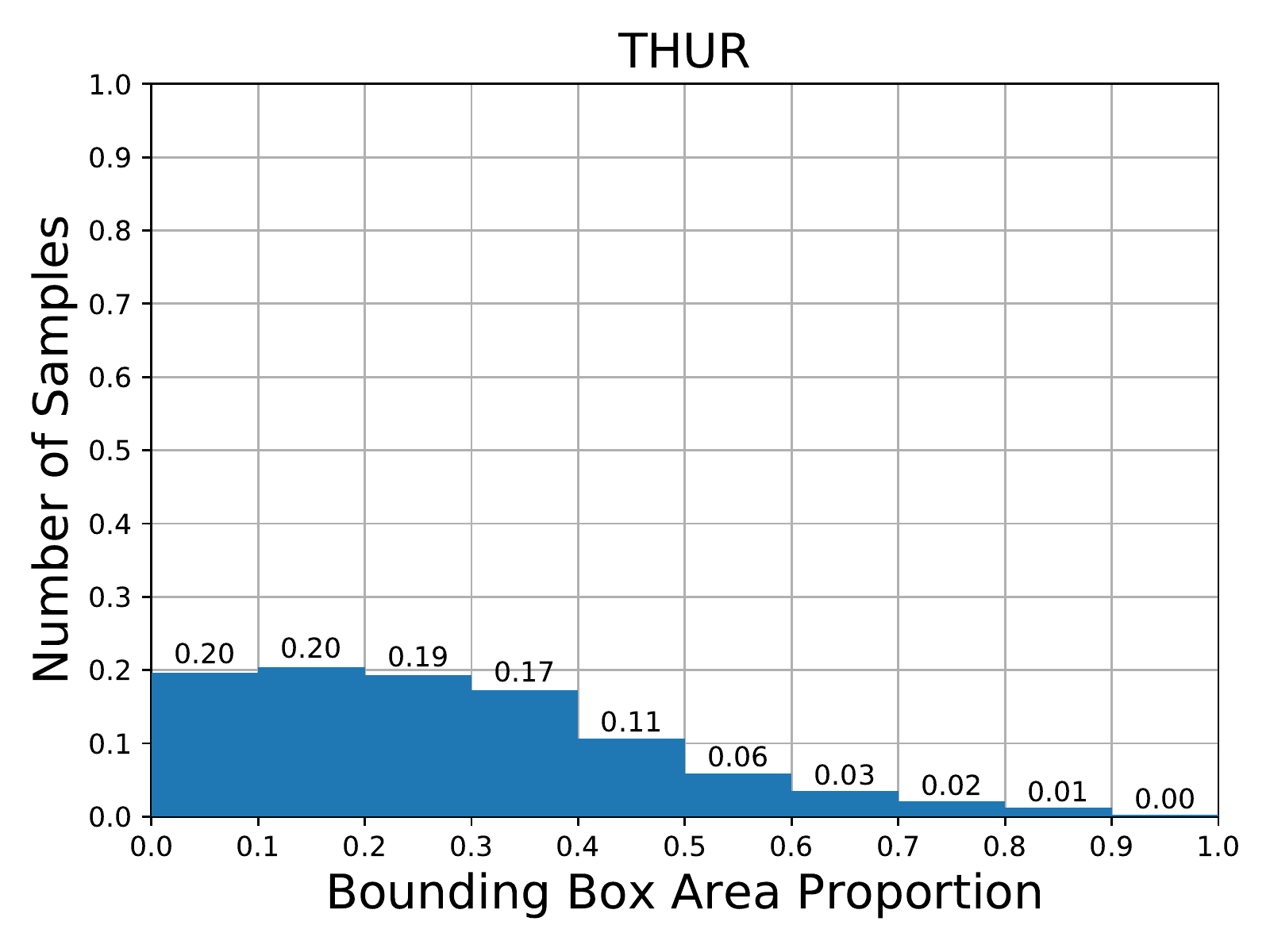}}

\subfloat{\includegraphics[width=0.35\linewidth]{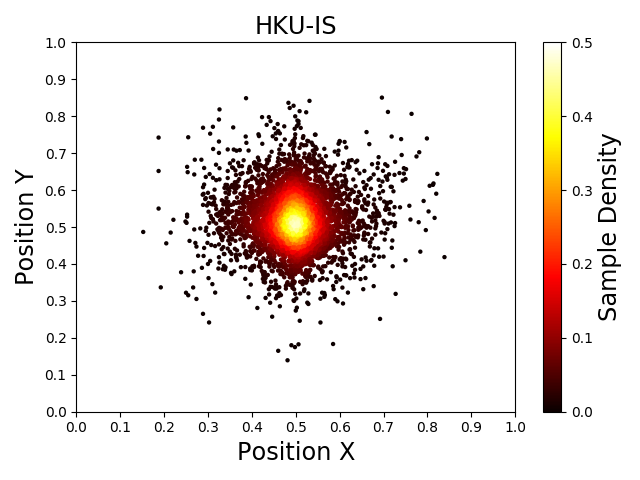}}
\hspace{0.5cm}
\subfloat{\includegraphics[width=0.35\linewidth]{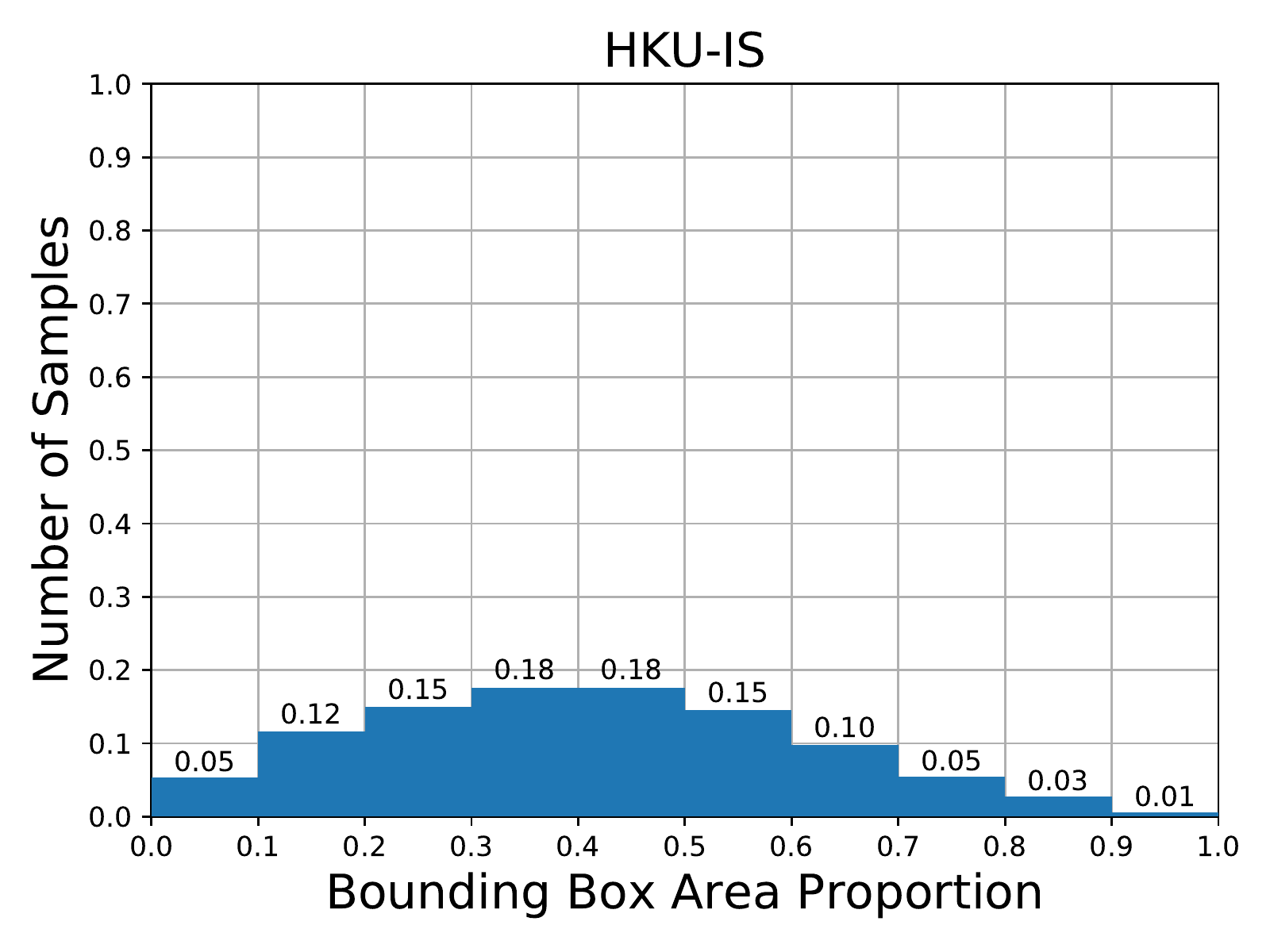}}

\subfloat{\includegraphics[width=0.35\linewidth]{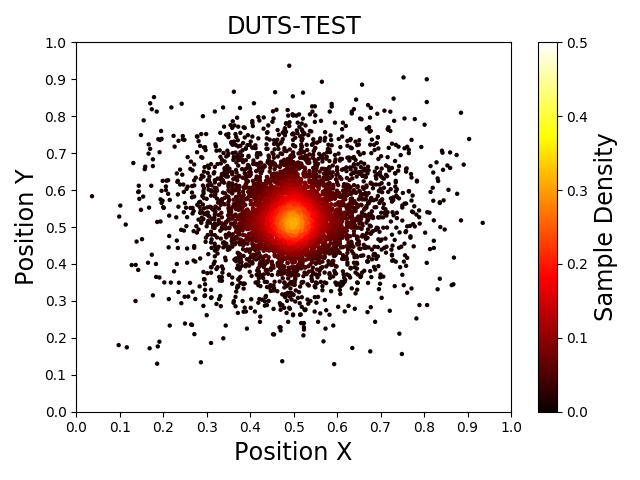}}
\hspace{0.5cm}
\subfloat{\includegraphics[width=0.35\linewidth]{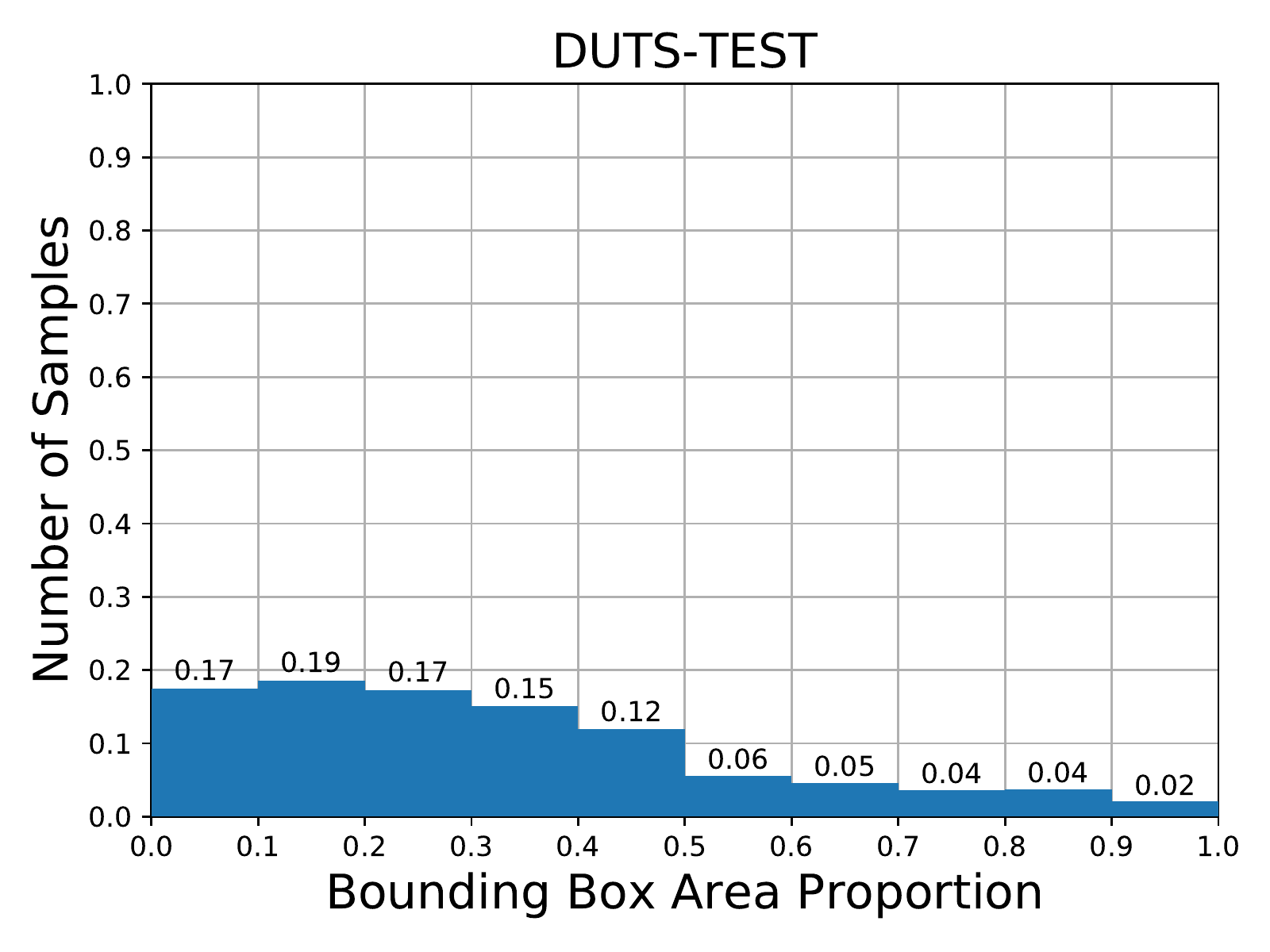}}

\subfloat{\includegraphics[width=0.35\linewidth]{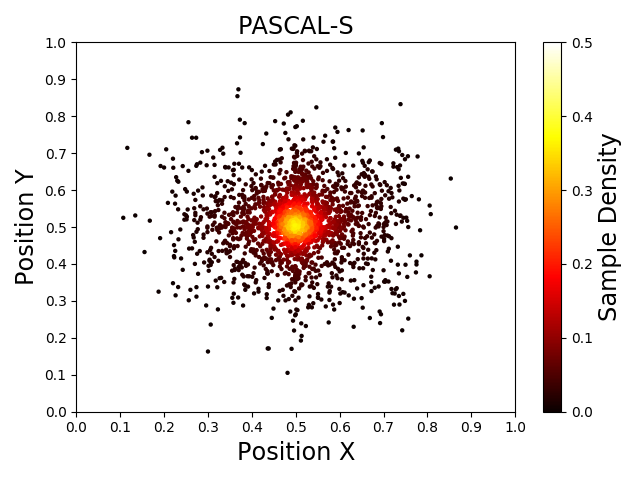}}
\hspace{0.5cm}
\subfloat{\includegraphics[width=0.35\linewidth]{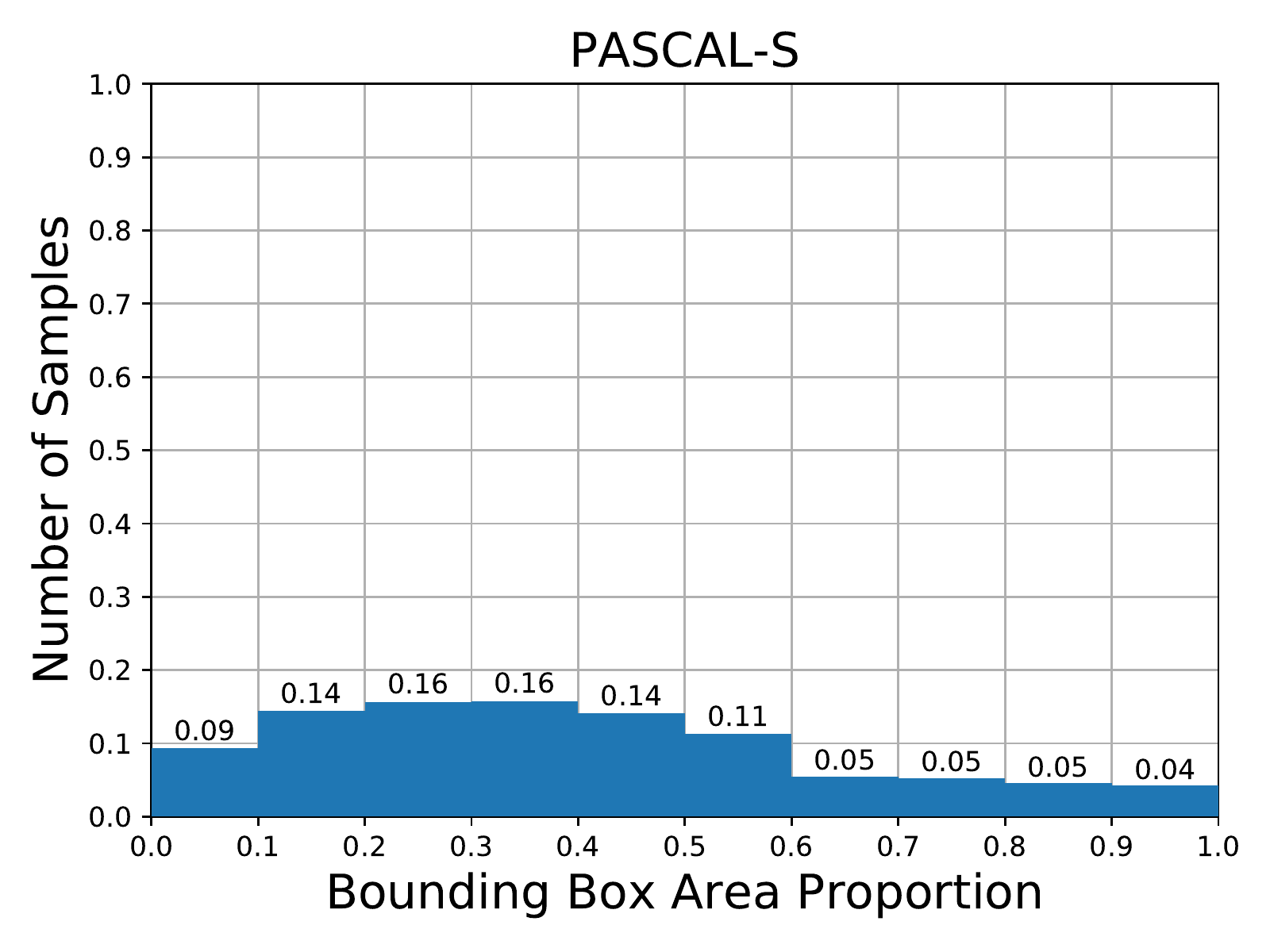}}

\subfloat{\includegraphics[width=0.35\linewidth]{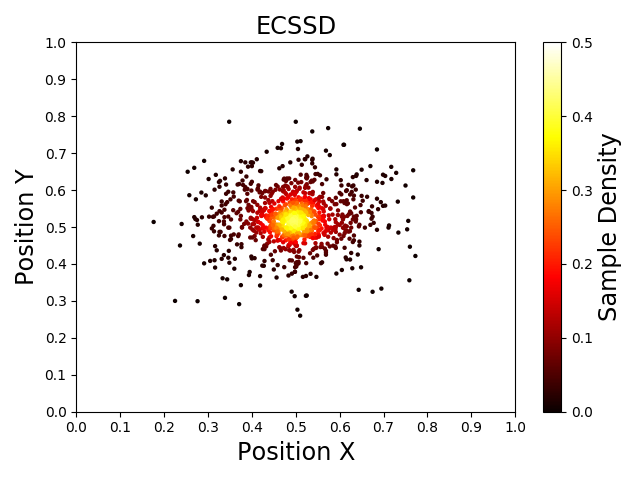}}
\hspace{0.5cm}
\subfloat{\includegraphics[width=0.35\linewidth]{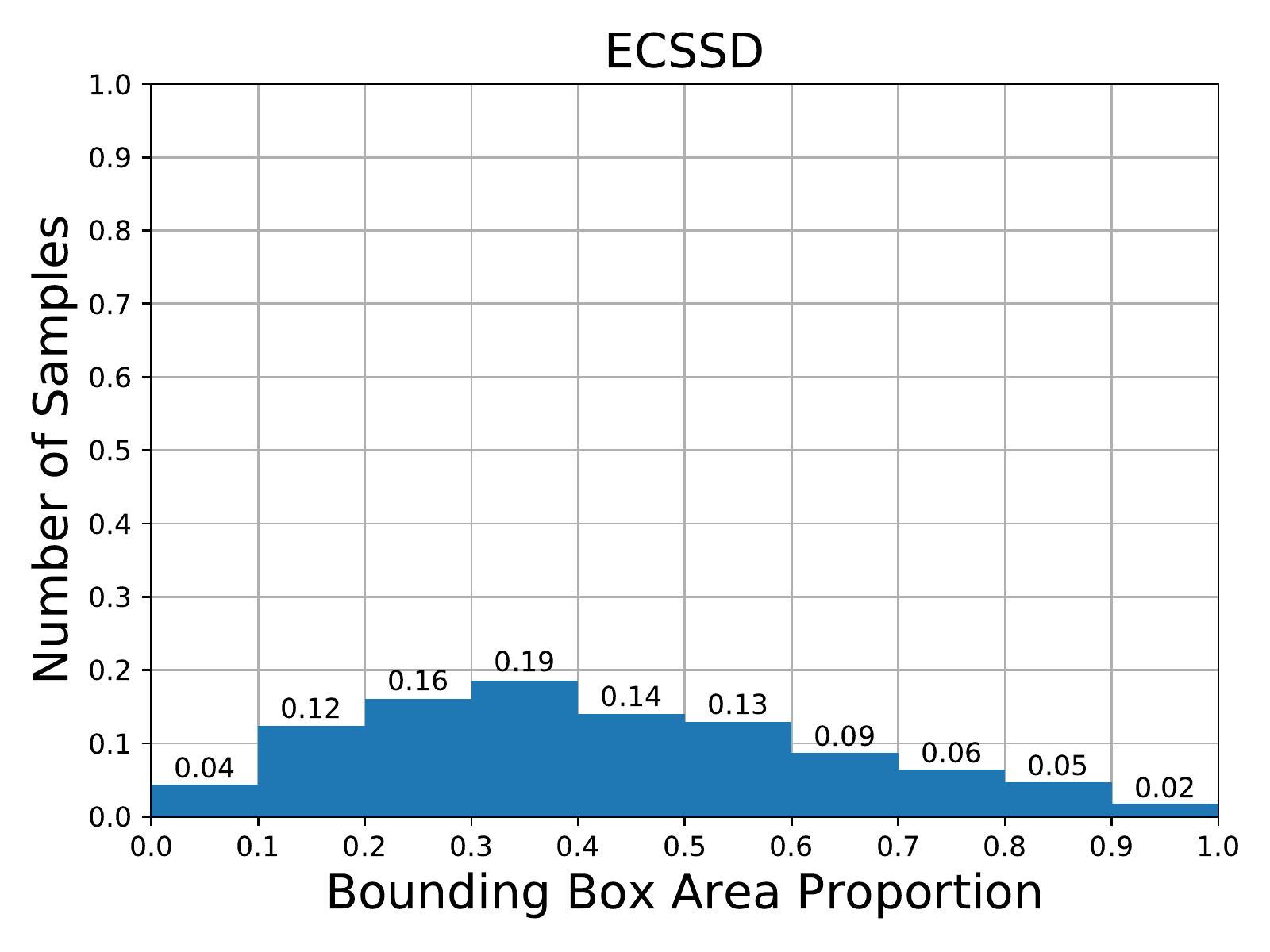}}

\caption{Position and size distribution per dataset for testing: DUT-OMRON, THUR, HKU-IS, DUTS-TEST, PASCAL-S, and ECSSD. The position distribution is demonstrated in a scatter plot of the normalized bounding box center coordinates. Additionally, a heat colormap represents the sample position density. For size, the bounding box area divided by image area is displayed in a 10-bin histogram.}
\label{fig:dist2}       
\end{figure}

When analyzing the salient datasets available in the literature, we noticed a large difference regarding the salient objects' size and position on the images of training and testing datasets. Fig.~\ref{fig:dist1} and Fig.~\ref{fig:dist2} show the position (left) and size (right) distribution of several dataset utilized in our work.

The position distribution shows each object bounding box's central location in a given dataset, while the size distribution shows the proportion of each salient object to the image size. In Fig.~\ref{fig:dist1}, the first row presents the position and size of the DUTS~\cite{wang2017duts} training set, utilized in our work as a baseline training dataset, and in the second row, the DUTS training set combined with our proposal augmentation, respectively. Fig.~\ref{fig:dist2} presents the position and size distribution of testing datasets DUT-OMRON~\cite{yang2013dutomron}, THUR15K~\cite{cheng2014thur},  HKU-IS~\cite{li2015hkuis}, DUTS-TEST~\cite{wang2017duts}, PASCAL-S~\cite{li2014pascals}, and ECSSD~\cite{shi2016ecssd}, respectively.

As presented on Fig.~\ref{fig:dist1}, the salient objects in the DUTS dataset, utilized for training, are often located in the center of the image and take between 20 and 60 percent of the image, while the salient objects in other datasets are more widespread in the image. Also, datasets like {DUT-OMRON}, THUR, {PASCAL-S}, and {DUTS-TEST}~(Fig.~\ref{fig:dist2}), have a higher occurrence of small objects. We generated more examples of small salient objects with different positions in the image with our method, which made the training dataset less biased.

The training dataset is composed of $k$ images. The index $i$ represents the $i$th image inside the training dataset. Each image $i$ has an object $o$, a background $b$, and a bounding box $\phi$ defined by two corner points $p=(x_{\min},y_{\min})=(x,y)$ and $p^\prime=(x_{\max},y_{\max})=(x^{\prime},y^{\prime})$. The bounding box has a width $w_\phi=x^\prime-x$, a height $h_\phi=y^\prime-y$, and an area $a_\phi=w_\phi \times h_\phi$. The background has a width $w_b$, a height $h_b$, and an area $a_b=w_b \times h_b$.

Aiming to resize the object $o$, we multiply its width and height by a scale factor $s_i$ as in Equation~\ref{eq:s}. If the resized object $o^\prime$ with $w_o^{\prime}=w_o \times s_i$ and $h_o^{\prime}=h_o \times s_i$ can fit $b$, then we proceed to the next step.

\begin{equation}
s_i = \sqrt{ \frac{f(i \bmod 3) \times a_b}{a_\phi} }
\label{eq:s}
\end{equation}

Let $R$ be a random variable following a uniform distribution with its range is defined by the function $f(i \bmod 3)$ as in Equation~\ref{eq:scale}. Those ranges were chosen to encourage a reduction in size since there are fewer samples with small objects on the DUTS training dataset.

\begin{equation}
    f(i \bmod 3) = 
    \begin{cases}
    R\in[0.075,0.1), i \bmod 3=0;\\
    R\in[0.1,0.2), i \bmod 3=1;\\
    R\in[0.2,0.3), i \bmod 3=2.
    \end{cases}
    \label{eq:scale}
\end{equation}

If the resized object $o^\prime$ cannot fit $b$, then there are two cases. In the first case, both $\phi$ and $b$ have the same orientation, landscape, or portrait, then $s$ is defined as in Equation~\ref{eq:s2}.

\begin{equation}
  s_i = 0.5 \times \min \left( \frac{w_b}{w_\phi},\frac{h_b}{h_\phi} \right) 
  \label{eq:s2}
\end{equation}

The scalar $s_i$ receives half of the highest possible value to resize the object without deforming it or causing the object to overflow the background. The second case, $\phi$, and $b$ have a different orientation, then $o$ is rotated by $-90^{\circ}$ degrees or $90^{\circ}$ degrees, the sign of the value is chosen randomly, then the same resize as the first case occurs. 

Ruiz \textit{et al.}~\cite{ruiz2019anda} proposed a random uniform translation on the resized object $o^\prime$ to diversify the position distribution. While this approach can lead to widespread in position distribution it have no way of predicting if the new location preserve salience. Instead of a random uniform translation we propose an intra-image optimization as follows: given the resized object $o^\prime$ and the background $b$ compute the feature vector $A^{*}$ of $o^\prime$ as described in  Section~\ref{sec:linearObjBg}. Slice $b$ in $o^\prime$ sized patches $b_{uv}$, $u$ and $v$ are the patches indexes. Regarding the margins, ensure that the patch still has $h_{o}^\prime \times w_{o}^\prime$. Then, compute for every patch $b_{uv}$ a feature vector $B_{uv}$ also the same way as described in Section~\ref{sec:linearObjBg}. Finally, find the maximal distance $d_c(A^{*},B_{uv})$, the $uv$ coordinate that maximizes the distance is the patch that $o^\prime$ will superimpose. Fig.~\ref{fig:flowchart} illustrates the intra-image optimization.


\section{Experiments}\label{sec:exp}

The majority of our experiments were performed in a single state-of-the-art neural network, the PoolNet Res2Net-50. For a fair comparison, we also present the results with the standard PoolNet ResNet-50.


\subsection{Datasets}

In our work, we used seven \gls*{sod} datasets widely used in the literature: DUTS~\cite{wang2017duts}, DUT-OMRON~\cite{yang2013dutomron},  THUR15K~\cite{cheng2014thur}, HKU-IS~\cite{li2015hkuis}, ECSSD~\cite{shi2016ecssd}, PASCAL-S~\cite{li2014pascals}, and SOC\cite{fan2018SOC}. DUTS is one of the largest datasets available on \gls*{sod} literature, with 10,553 training images and 5,017 testing images. DUT-OMRON is a large dataset, being composed of 5,168 images with a great variety of objects in challenging backgrounds. THUR15K has 6,232 labeled images divided into five categories: butterfly, coffee mug, dog jump, giraffe, and plane. HKU-IS is composed of 4,447 images with low contrast and multi salient objects. ECSSD has 1,000 images with a high content variety and different objects. PASCAL-S derives from the validation set of the PASCAL VOC 2012~\cite{pascal-voc-2012} dataset and is composed of 2,205 images. Salient Objects in Clutter~(SOC) is a challenging dataset that, differently from other \gls*{sod} datasets, includes images with no salient objects and includes salient objects with real-world challenging visual occurrences like motion blur, occlusion, and cluttered background.


\subsection{Metrics}

To evaluate our data augmentation technique's impact, we use five evaluation metrics: Precision, Recall, F-score,  \gls*{mae}, and the Structure-measure~\cite{fan2017smeasure}.

In the F-score, we use $\beta$ equals to $0.3$ as done in ~\cite{liu2019poolnet,ruiz2019anda}, in order to weight precision more than recall. It is important to state that these metrics are designed to compare two binary maps and the salience maps evaluated are non-binary. For that reason, binarization is necessary for us to use those metrics. The binarization process is threshold-dependent, so we follow the evaluation procedure used on~\cite{liu2019poolnet} that defines $F_{\beta}^{*}$~(Equation~\ref{eq:bestfscore}), $P^{*}$~(Equation~\ref{eq:bestprecision}), $R^{*}$~(Equation~\ref{eq:bestrecall}).

\begin{equation}
    P_{\mu}(th) = \dfrac{1}{N}\times\sum^{N}_{i=1}P(th,i)
    \label{eq:meanprecision}
\end{equation}

\begin{equation}
    R_{\mu}(th) = \dfrac{1}{N}\times\sum^{N}_{i=1}R(th,i)
    \label{eq:meanrecall}
\end{equation}

\begin{equation}
    F_{\beta}^{*} = max(\{ \dfrac{(1+\beta^2) \times P_{\mu}(th) \times R_{\mu}(th)}{\beta^2 \times P_{\mu}(th) + R_{\mu}(th)} | 0<th<255\})
    \label{eq:bestfscore}
\end{equation}

$N$ is the number of images, $P(th,i)$ and $R(th,i)$ are the Precision of an image binarized $i$ with threshold $th$, and the Recall with the same parameters, respectively. Finally, $bestTh$ is the threshold $th$ that produces the maximal $F_{\beta}^{*}$.

\begin{equation}
    P^{*} = P_{\mu}(bestTh)
    \label{eq:bestprecision}
\end{equation}

\begin{equation}
    R^{*} = R_{\mu}(bestTh)
    \label{eq:bestrecall}
\end{equation}

Structure-measure, or simply s-measure, proposed on~\cite{fan2017smeasure}, evaluate non-binary foreground maps. This metric simultaneously evaluates region-aware and object-aware structural similarity between a Salience map and a Ground Truth map.


\subsection{Quantitative Results}

We present our findings in this subsection after training the PoolNet Res2Net50 neural network with different data augmentation techniques.  Table~\ref{tab:tests} shows the comparison between our baseline architecture, PoolNet with ResNet-50 as a backbone, trained with one data augmentation technique, the PoolNet with Res2Net-50 as the backbone, trained with no data augmentation; and other six training sets, each with a different data augmentation technique. 

F-measure, Precision, Recall were not computed for the SOC dataset, as was recommended by Fan et al.~\cite{fan2018SOC}, since SOC contains many non-salient images, for such, the ground-truth is an all-zero matrix, thus directly using the F-measure may result in an inaccurate score.

Flip operations are almost always used as data augmentation due it is simplicity and effectiveness~\cite{GuoSymmetry2018,liu2019poolnet,wei2019f3net,wang2019progressive}. So, here H-Flip was chosen to represent how our method compares with one that uses affine transformations only and to ensure a fair comparison with results presented by~\cite{liu2019poolnet}. GridMask was chosen to represent how a data erasure method compares with ours. To measure the impact of the newly proposed improvements, we also present a comparison with our previous work ANDA~\cite{ruiz2019anda}.


\begin{table}[!htb]
\caption{Comparison between baseline and data augmented results. The best $F_{\beta}^{*}$ is highlighted in \textcolor{blue}{\textbf{bold blue text}}. The best S-score is highlighted in \textbf{bold text}. DUT-O* is an abbreviation of DUT-OMRON. DUTS-TE* is an abbreviation of DUTS-TEST. }

\resizebox{\linewidth}{!}{
\begin{tabular}{ccccccccccccc}

\toprule
\multicolumn{1}{c}{\textbf{Experiment}} &
\multicolumn{1}{c}{\textbf{Metric}} &
\multicolumn{1}{c}{\textbf{DUT-O*}} &
\multicolumn{1}{c}{\textbf{THUR15K}}& 
\multicolumn{1}{c}{\textbf{PASCAL-S}} &
\multicolumn{1}{c}{\textbf{DUTS-TE*}}&
\multicolumn{1}{c}{\textbf{HKU-IS}}&
\multicolumn{1}{c}{\textbf{ECSSD}}&
\multicolumn{1}{c}{\textbf{SOC}}
\\ \midrule

\multicolumn{1}{c}{} &
\multicolumn{1}{c}{\textbf{N\# Images}} &
\multicolumn{1}{c}{\textbf{5,168}} &
\multicolumn{1}{c}{\textbf{6,232}} & 
\multicolumn{1}{c}{\textbf{2,205}} &
\multicolumn{1}{c}{\textbf{5,017}} &
\multicolumn{1}{c}{\textbf{4,447}} &
\multicolumn{1}{c}{\textbf{1,000}} &
\multicolumn{1}{c}{\textbf{6,000}}
\\ \midrule

\makecell{PoolNet\\ \gls*{resnet}-50\\Baseline\\H-Flip Only\\\cite{liu2019poolnet}} & 
\makecell{S-score$\uparrow$ \\ $F_{\beta}^{*}$$\uparrow$ \\ $P^{*}$$\uparrow$ \\ $R^{*}$$\uparrow$ \\ \gls*{mae}$\downarrow$} &
\makecell{0.8341\\0.8305 \\0.8602 \\0.7448 \\0.0554} &
\makecell{0.8347\\0.8026 \\0.7945 \\0.8309 \\0.0698} &
\makecell{0.8368\\0.8716 \\0.9058 \\0.7744 \\0.0733} &
\makecell{0.8824\\0.8858 \\0.9104 \\0.8128 \\0.0397} &
\makecell{0.9152\\0.9343 \\0.9583 \\0.8624 \\0.0327} &
\makecell{0.9207\\0.9444 \\0.9649 \\0.8821 \\0.0388}&
\makecell{0.8464\\-\\-\\-\\0.0913}

\\ \midrule

\makecell{No data aug} & 
\makecell{S-score$\uparrow$ \\$F_{\beta}^{*}$$\uparrow$ \\ $P^{*}$$\uparrow$ \\ $R^{*}$$\uparrow$ \\ \gls*{mae}$\downarrow$} &
\makecell{0.8219\\0.8391 \\ 0.8872 \\ 0.7106 \\ 0.0513} &
\makecell{0.8369\\0.8092 \\ 0.8023 \\ 0.8331 \\ 0.0650} & 
\makecell{0.8313\\0.8714 \\ 0.9086 \\ 0.7668 \\ 0.0726}&
\makecell{0.8760\\0.8880 \\ 0.9197 \\ 0.7964 \\ 0.0376}&
\makecell{0.9087\\0.9318 \\ 0.9585 \\ 0.8526 \\ 0.0321}&
\makecell{0.9202\\0.9450 \\0.9670 \\0.8785 \\0.0365}&
\makecell{0.8623\\-\\-\\-\\0.0727}
\\ \addlinespace[1.2mm]

\makecell{GridMask\\Only}&
\makecell{S-score$\uparrow$ \\$F_{\beta}^{*}$$\uparrow$ \\ $P^{*}$$\uparrow$ \\ $R^{*}$$\uparrow$ \\ \gls*{mae}$\downarrow$} &

\makecell{0.8271\\0.8333\\0.8709\\0.7285\\ 0.0526}&
\makecell{0.8341\\0.8040\\0.7957\\0.8332\\0.0684}&
\makecell{0.8359\\0.8700\\0.9034\\0.7744\\0.0717}&
\makecell{0.8774\\0.8848\\0.9141\\0.7993\\0.0388}&
\makecell{0.9112\\0.9327\\0.9574\\0.8588\\0.0321}&
\makecell{0.9232\\0.9471\\0.9650\\0.8917\\0.0355}&
\makecell{0.8476\\-\\-\\-\\0.0881}
\\ \addlinespace[1.2mm]

\makecell{ANDA Only\\~\cite{ruiz2019anda}}&
\makecell{S-score$\uparrow$ \\$F_{\beta}^{*}$$\uparrow$ \\ $P^{*}$$\uparrow$ \\ $R^{*}$$\uparrow$ \\ \gls*{mae}$\downarrow$} &

\makecell{0.8337\\0.8289 \\ 0.8544 \\ 0.7539 \\ 0.0572} 
&\makecell{0.8355\\0.8048 \\ 0.7912 \\ 0.8539 \\ 0.0692} &
\makecell{0.8372\\0.8686 \\ 0.9013 \\ 0.7750 \\ 0.0727}&
\makecell{0.8801\\0.8845 \\ 0.9103 \\ 0.8081 \\ 0.0404}&
\makecell{0.9161\\0.9346\\0.9571 \\0.8668\\ 0.0305}&
\makecell{0.9242\\0.9463\\0.9650\\0.8890\\0.0352}&
\makecell{0.8515\\-\\-\\-\\0.0851}
\\ \addlinespace[1.2mm]

\makecell{IDA Only}&
\makecell{S-score$\uparrow$ \\$F_{\beta}^{*}$$\uparrow$ \\ $P^{*}$$\uparrow$ \\ $R^{*}$$\uparrow$ \\ \gls*{mae}$\downarrow$} &

\makecell{0.8362\\0.8371\\0.8723\\0.7376\\0.0538}&
\makecell{0.8393\\0.8102\\0.8018\\0.8397\\0.0667}&
\makecell{0.8351\\0.8714\\0.9068\\0.7710\\0.0722}&
\makecell{0.8831\\0.8870\\0.9108\\0.8159\\0.0387}&
\makecell{0.9144\\0.9336\\0.9565\\0.8645\\0.0314}&
\makecell{0.9244\\\textbf{\textcolor{blue}{0.9475}}\\0.9666\\0.8891\\0.0363}&
\makecell{0.8623\\-\\-\\-\\0.0748}
\\ \addlinespace[1.2mm]

\makecell{H-Flip Only}&
\makecell{S-score$\uparrow$ \\$F_{\beta}^{*}$$\uparrow$ \\ $P^{*}$$\uparrow$ \\ $R^{*}$$\uparrow$ \\ \gls*{mae}$\downarrow$} &

\makecell{0.8350\\0.8392 \\ 0.8791 \\ 0.7292 \\ 0.0518} 
&\makecell{0.8439\\0.8133 \\ 0.8025 \\ 0.8517 \\ 0.0649} &
\makecell{0.8376\\0.8735 \\ 0.9067 \\ 0.7784 \\ 0.0719}&
\makecell{0.8850\\0.8938 \\ 0.9194 \\ 0.8178 \\ 0.0371}&
\makecell{0.9152\\0.9369 \\ 0.9625 \\ 0.8607 \\ 0.0312}&
\makecell{0.9240\\0.9471\\0.9670 \\0.8863 \\0.0365}&
\makecell{0.8651\\-\\-\\-\\0.0729}

\\ \addlinespace[1.2mm]

\makecell{GridMask + \\H-Flip}&
\makecell{S-score$\uparrow$ \\$F_{\beta}^{*}$$\uparrow$ \\ $P^{*}$$\uparrow$ \\ $R^{*}$$\uparrow$ \\ \gls*{mae}$\downarrow$} &
\makecell{0.8369\\0.8398 \\ 0.8843 \\ 0.7193 \\ 0.0555} &
\makecell{0.8415\\0.8135 \\ 0.8086 \\ 0.8300 \\ 0.0682} &
\makecell{\textbf{0.8391}\\0.8730 \\ 0.9150 \\ 0.7571 \\ 0.0743}&
\makecell{0.8862\\0.8930 \\ 0.9226 \\ 0.8066 \\ 0.0396}&
\makecell{0.9156\\0.9345 \\ 0.9626 \\ 0.8517 \\ 0.0340}&
\makecell{0.9229\\0.9459 \\0.9651 \\0.8869 \\0.0404}&
\makecell{0.8472\\-\\-\\-\\0.0932}

\\ \addlinespace[1.2mm]

\makecell{IDA +\\GridMask + \\H-Flip}&
\makecell{S-score$\uparrow$ \\$F_{\beta}^{*}$$\uparrow$ \\ $P^{*}$$\uparrow$ \\ $R^{*}$$\uparrow$ \\ \gls*{mae}$\downarrow$} &
\makecell{\textbf{0.8441}\\\textbf{\textcolor{blue}{0.8444}} \\ 0.8910 \\ 0.7191 \\ 0.0540} &
\makecell{\textbf{0.8449}\\\textbf{\textcolor{blue}{0.8167}} \\ 0.8130 \\ 0.8294 \\ 0.0663} &
\makecell{0.8382\\\textbf{\textcolor{blue}{0.8762}} \\ 0.9145 \\ 0.7690 \\ 0.0737}&
\makecell{\textbf{0.8918}\\\textbf{\textcolor{blue}{0.8989}} \\0.9268 \\ 0.8170 \\ 0.0377}&
\makecell{\textbf{0.9190}\\\textbf{\textcolor{blue}{0.9386}} \\0.9641 \\ 0.8626 \\ 0.0324}&
\makecell{\textbf{0.9256}\\0.9468 \\0.9640 \\0.8936 \\0.0388}&
\makecell{\textbf{0.8711}\\-\\-\\-\\0.0710}

\\ \bottomrule

\end{tabular}
}
\label{tab:tests}

\end{table}

We concluded that the drastic change of backbone architecture only improves the overall results by a small margin, i.e., for F-measure a gain of 1.07 on the best case and 0.19 on the worst-case comparing the ResNet-50 (baseline with H-flip) with the Res2Net-50 H-flip only.

After the backbone comparison, we analyzed how much the data augmentation can further improve those gains. Our method alone (IDA Only on Table~\ref{tab:tests}) was not enough to surpass the traditional H-Flip but did surpass the ANDA only and GridMask Only in F-measure in five of six datasets. We also experimented with a combination of techniques.

Maintaining the same architecture (Res2Net-50), we further improved the results by 0.52 on the best case, -0.03 in the worst case, comparing Res2Net50 H-flip only to IDA (Ours) + GridMask + H-Flip, this improvement was made by only applying data augmentation techniques surpassing the previous results on these datasets~\cite{liu2019poolnet,gao2019res2net}.

The combination of our proposed method IDA with GridMask and H-Flip achieved the best s-measure in all evaluated datasets (except in the PASCAL-S), and best F-score in all datasets except the ECSSD in which the IDA only achieved the best F-score. In the case of precision, it achieved the best result in four of the six datasets. 


\subsection{Ranking models}

To rank the performance of the saliency models  
we use an average ranking technique. Similarly to ~\cite{contreras2019genetic}, this approach can summarize multiple metrics into a single, more readable value. The set of test images were generated by combining all testing datasets except SOC, since it contains images without any salience, resulting in a total of $24,069$ images. Nine metrics were chosen to compute the ranking of each model:

\begin{enumerate}[leftmargin=1cm,align=left]
    \item S-measure
    \item $F_{\beta}$
    \item Precision~(Pr)
    \item Recall~(Re)
    \item MAE
    \item Specificity~(Sp)
    \item False Positive Rate~(FPR)
    \item False Negative Rate~(FNR)
    \item Percentage of Wrong Classifications~(PWC)
\end{enumerate}

The average ranking of a saliency model $M_i$ is described by Equation~\ref{eq:rank}. $\tau$ is a set of images in a test dataset, and $P$ is the number of metrics utilized to evaluate the model $M$. To ensure a fair comparison using binary metrics, we use the same fixed threshold for every model.

\begin{equation}
    R_i = \dfrac{1}{P}\sum_{j=1}^{P}rank(metric_j(M_i(\tau));metric_j(M_k(\tau)), \forall k \neq i ) 
    \label{eq:rank}
\end{equation}

For metrics which higher values are desirable a descending sort is employed, e. g., using a $metric_j$ the models $M_0, M_1, M_2$ achieved the values $[0.97,0.94,0.95]$ respectively, so $M_0$ is ranked as $1$, $M_2$ as $2$, $M_1$ as $3$. An ascending order is employed otherwise. In both cases, models with rank close to one are considered better.

As presented in Table~\ref{tab:tests}, the IDA, when combined with other techniques, achieved the best results in the majority of test datasets. So, in this section, we present further variations of training and the results, which are presented in Table~\ref{ranks}. Three variations of the {(IDA + GridMask + H-flip)} approach were evaluated, and are described as follows:

\begin{itemize}
    \item In (IDA+Grid+H-flip)* we use a proportion of 2 original images to 1 IDA image, resulting in training 31,659 images; this makes the synthetic images have a smaller impact on the training. 
    \item  (IDA+Grid+H-flip)$^\prime$  stands for a reduction of samples generated from IDA; the 611 top worst f-measure training samples were removed from the training set, resulting in 20,495 images. The evaluation to find the worst samples was done using a pre-trained model.  
    \item  (IDA+Grid+H-flip)$^{\prime\prime}$ stands for the same as (IDA+Grid+H-flip) with the difference that, instead of using a fixed probability of applying the GridMask in all steps, the probability increases accordingly to the epoch during training, it starts with a probability of 0.0 at the first epoch and ends with a probability of 1.0 at the final epoch.
\end{itemize}

\begin{table}[!htb]
\centering
\caption{Average ranking between the data augmentations techniques. Our method, combined with others~(IDA+Grid+H-flip) achieved the rank closer to one, getting ahead of the other methods. The test images were generated by combining all testing datasets except SOC, due to the images without any salience, with a total of 24,069 images.}

\begin{tabular}{lcc}
\toprule
Data Augmentation Technique & Average ranking \\ \midrule
IDA+Grid+H-flip       & 2.7 \\
(IDA+Grid+H-flip)$^\prime$   & 4.1 \\
IDA+H-flip          & 4.4 \\
\midrule
H-flip only           & 5.1 \\
\midrule
(IDA+Grid+H-flip)*    & 6.2 \\
Grid+H-flip           & 6.3 \\
(IDA+Grid+H-flip)$^{\prime\prime}$ & 6.3\\
IDA only              & 6.5 \\
No data Augmentation           & 7.7 \\ 
ANDA Only             & 8.1 \\ 
Grid Only             & 8.2 \\ 
\bottomrule 
\end{tabular}
\label{ranks}

\end{table}

The ranking procedure considers multiple metrics that can describe different aspects of the resulting segmentation masks and summarize them, making for a more readable result. Three different models containing our generated images were ranked above the standard horizontal flip and the GridMask method, meaning that the results were superior in multiple metrics. 

\subsection{Qualitative results}

\begin{figure}
\centering
\captionsetup[subfigure]{labelformat=empty}
\raisebox{0.0in}{\rotatebox{90}{\tiny{THUR15K}}}
\subfloat{\includegraphics[width=0.17\linewidth]{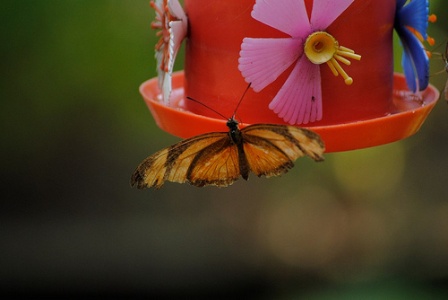}}
\hspace{0.1mm}
\subfloat{\includegraphics[width=0.17\linewidth]{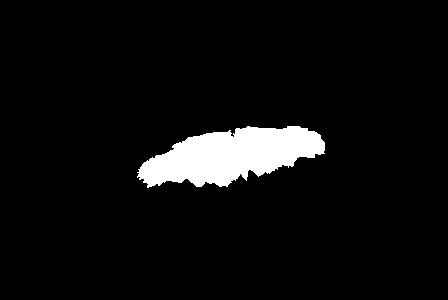}}
\hspace{0.1mm}
\subfloat{\includegraphics[width=0.17\linewidth]{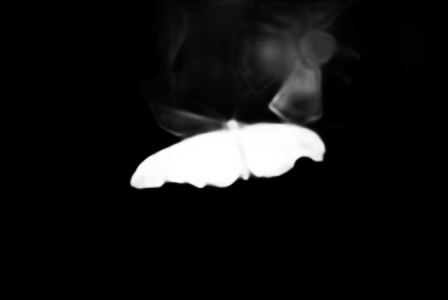}}
\hspace{0.1mm}
\subfloat{\includegraphics[width=0.17\linewidth]{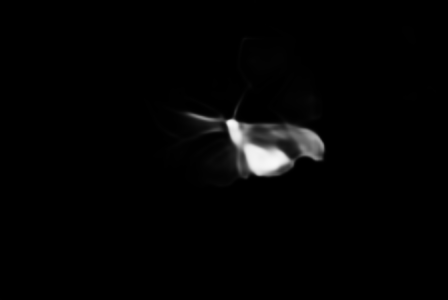}}
\hspace{0.1mm}
\subfloat{\includegraphics[width=0.17\linewidth]{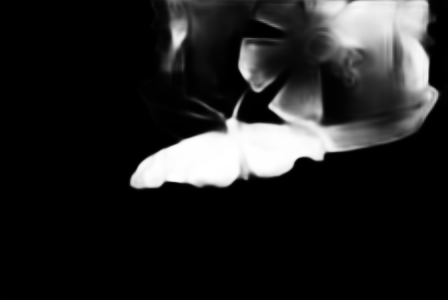}}

\vspace{1.0mm}
\raisebox{0.in}{\rotatebox{90}{\tiny{DUT-O*}}}
\subfloat{\includegraphics[width=0.17\linewidth]{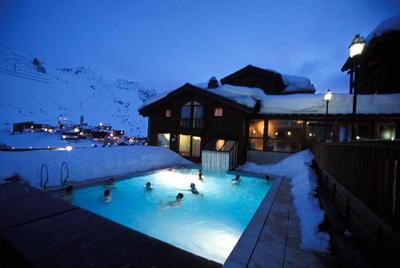}}
\hspace{0.1mm}
\subfloat{\includegraphics[width=0.17\linewidth]{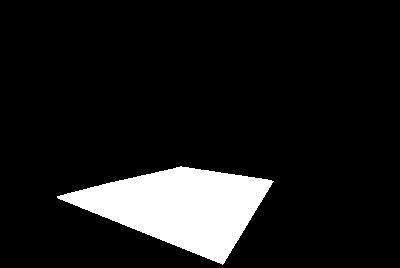}}
\hspace{0.1mm}
\subfloat{\includegraphics[width=0.17\linewidth]{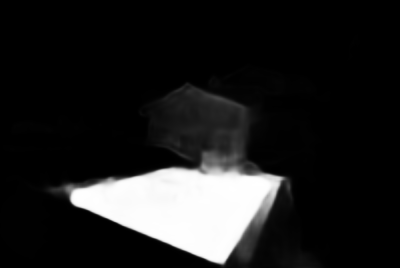}}
\hspace{0.1mm}
\subfloat{\includegraphics[width=0.17\linewidth]{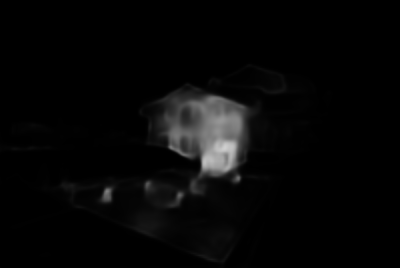}}
\hspace{0.1mm}
\subfloat{\includegraphics[width=0.17\linewidth]{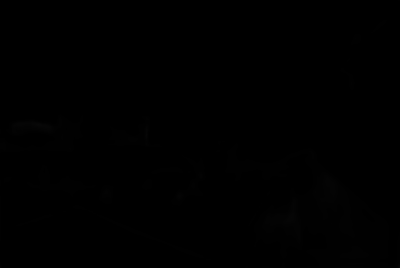}}

\vspace{1.0mm}
\raisebox{0.in}{\rotatebox{90}{\tiny{DUTS-TE*}}}
\subfloat{\includegraphics[width=0.17\linewidth]{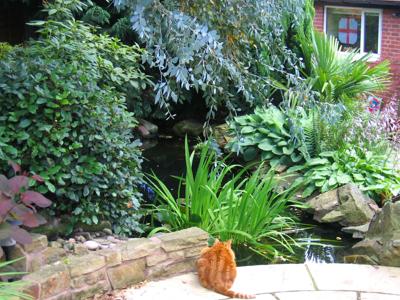}}
\hspace{0.1mm}
\subfloat{\includegraphics[width=0.17\linewidth]{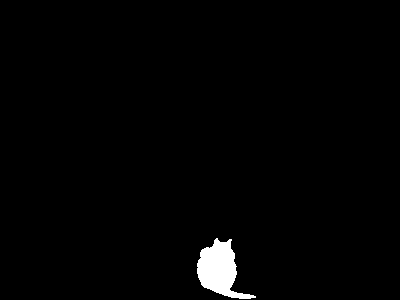}}
\hspace{0.1mm}
\subfloat{\includegraphics[width=0.17\linewidth]{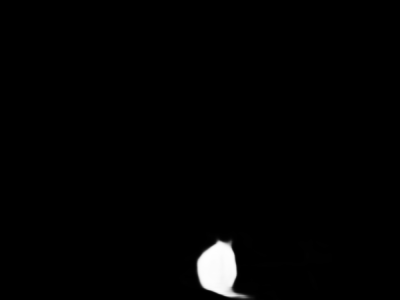}}
\hspace{0.1mm}
\subfloat{\includegraphics[width=0.17\linewidth]{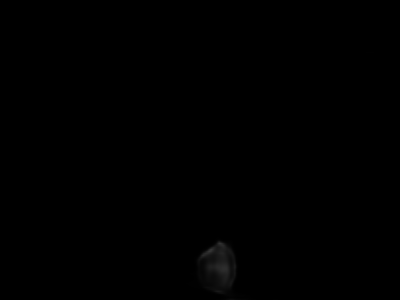}}
\hspace{0.1mm}
\subfloat{\includegraphics[width=0.17\linewidth]{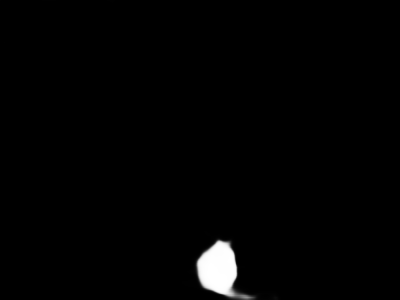}}

\vspace{1.0mm}
\raisebox{0.in}{\rotatebox{90}{\tiny{HKU-IS}}}
\subfloat{\includegraphics[width=0.17\linewidth]{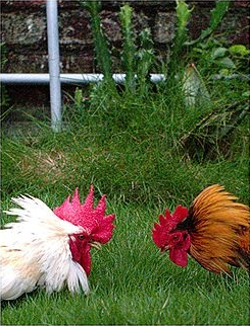}}
\hspace{0.1mm}
\subfloat{\includegraphics[width=0.17\linewidth]{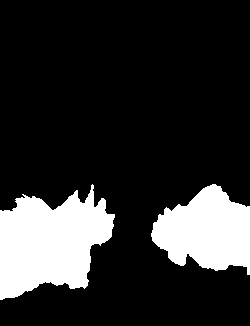}}
\hspace{0.1mm}
\subfloat{\includegraphics[width=0.17\linewidth]{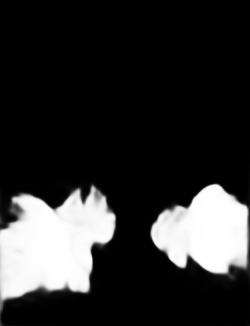}}
\hspace{0.1mm}
\subfloat{\includegraphics[width=0.17\linewidth]{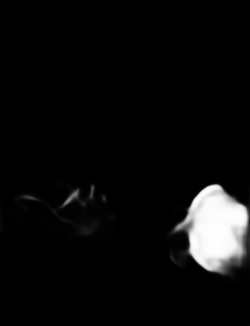}}
\hspace{0.1mm}
\subfloat{\includegraphics[width=0.17\linewidth]{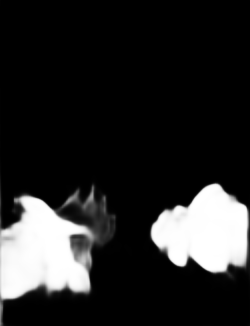}}

\vspace{1.0mm}
\raisebox{0.in}{\rotatebox[origin=lb]{90}{\tiny{ECSSD}}}
\subfloat{\includegraphics[width=0.17\linewidth]{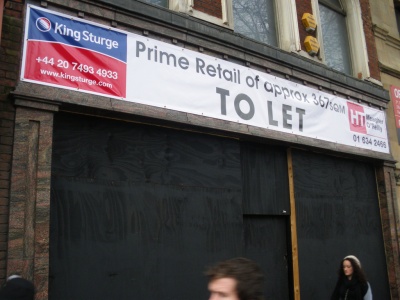}}
\hspace{0.1mm}
\subfloat{\includegraphics[width=0.17\linewidth]{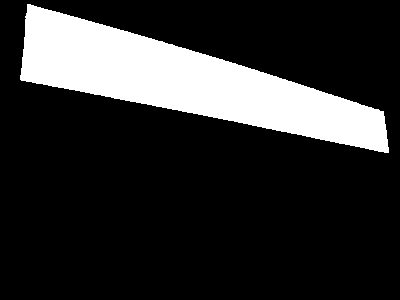}}
\hspace{0.1mm}
\subfloat{\includegraphics[width=0.17\linewidth]{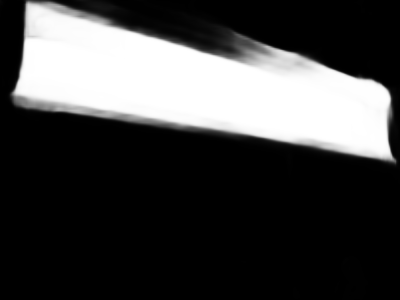}}
\hspace{0.1mm}
\subfloat{\includegraphics[width=0.17\linewidth]{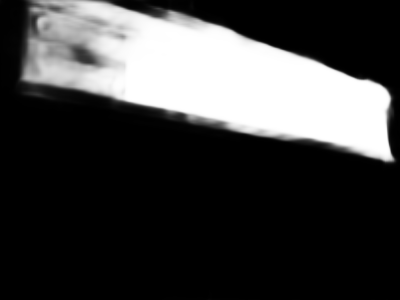}}
\hspace{0.1mm}
\subfloat{\includegraphics[width=0.17\linewidth]{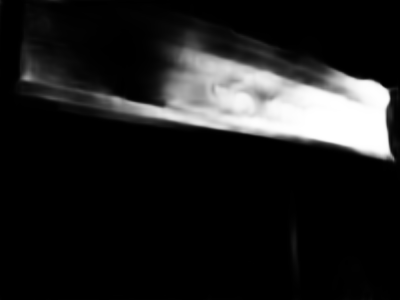}}

\vspace{1.0mm}
\setcounter{subfigure}{0}
\raisebox{0.in}{\rotatebox[origin=lb]{90}{\tiny{PASCAL-S}}} 
\subfloat{\includegraphics[width=0.17\linewidth]{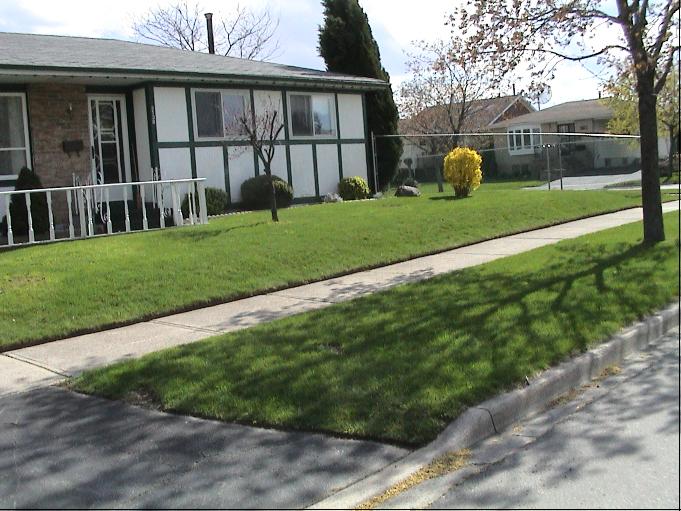}}
\hspace{0.1mm}
\subfloat{\includegraphics[width=0.17\linewidth]{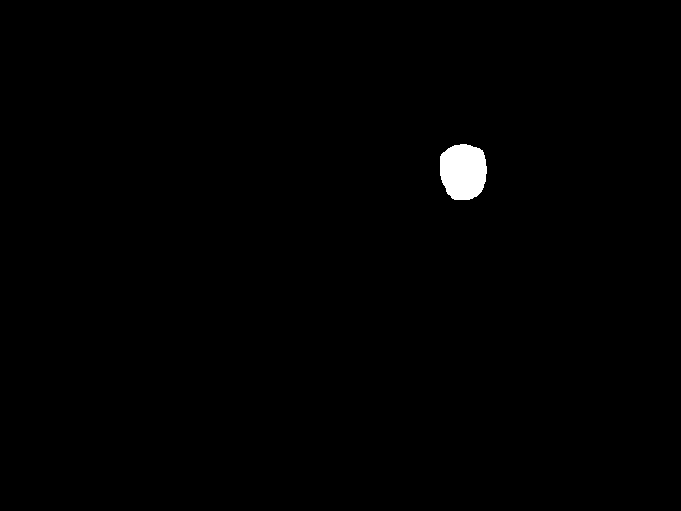}}
\hspace{0.1mm}
\subfloat{\includegraphics[width=0.17\linewidth]{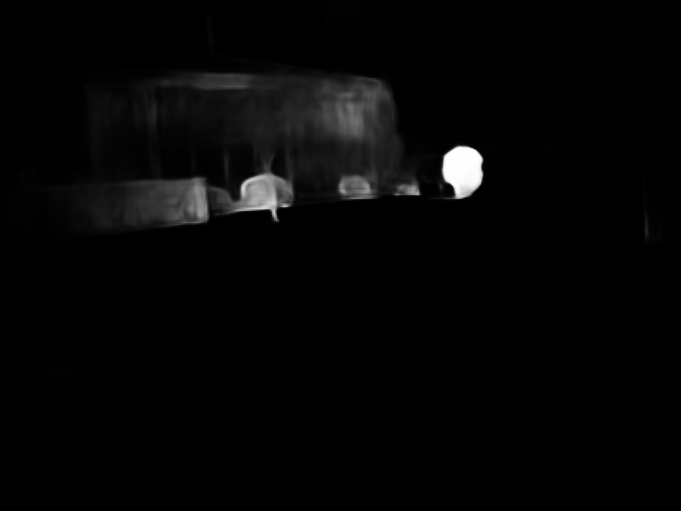}}
\hspace{0.1mm}
\subfloat{\includegraphics[width=0.17\linewidth]{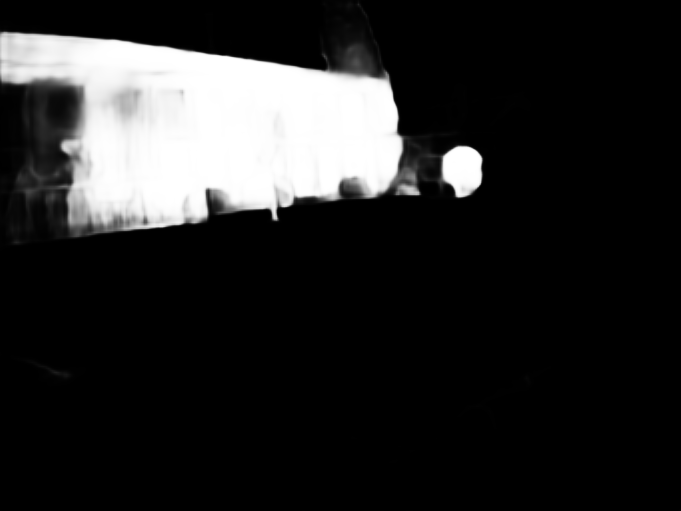}}
\hspace{0.1mm}
\subfloat{\includegraphics[width=0.17\linewidth]{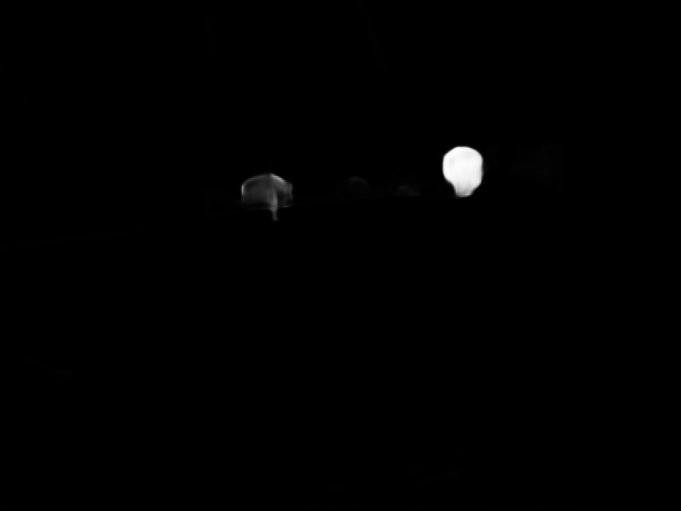}}

\vspace{1.0mm}
\setcounter{subfigure}{0}
\raisebox{0.in}{\rotatebox[origin=lb]{90}{\tiny{SOC}}} 
\subfloat[Image]{\includegraphics[width=0.17\linewidth]{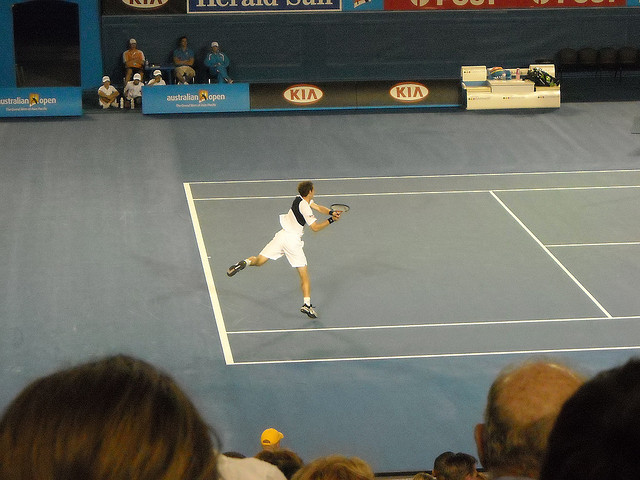}}
\hspace{0.1mm}
\subfloat[GT]{\includegraphics[width=0.17\linewidth]{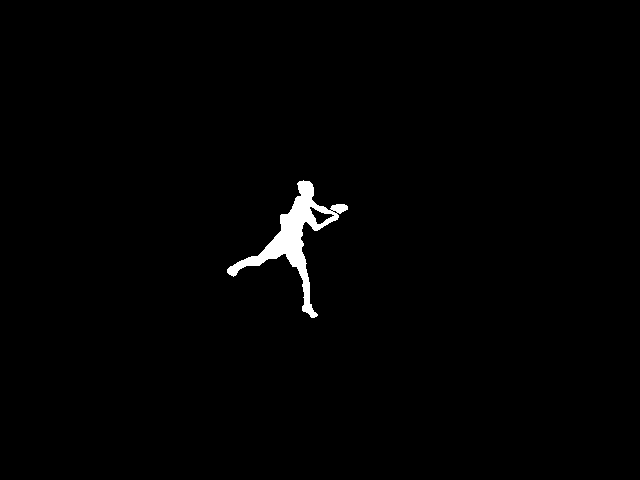}}
\hspace{0.1mm}
\subfloat[IDA*]{\includegraphics[width=0.17\linewidth]{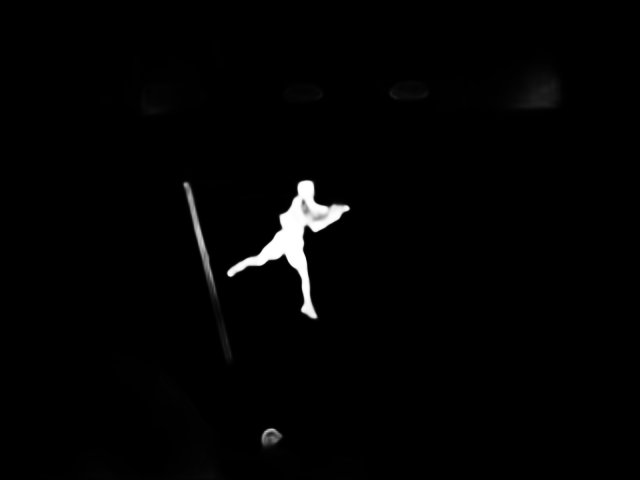}}
\hspace{0.1mm}
\subfloat[H-Flip Only]{\includegraphics[width=0.17\linewidth]{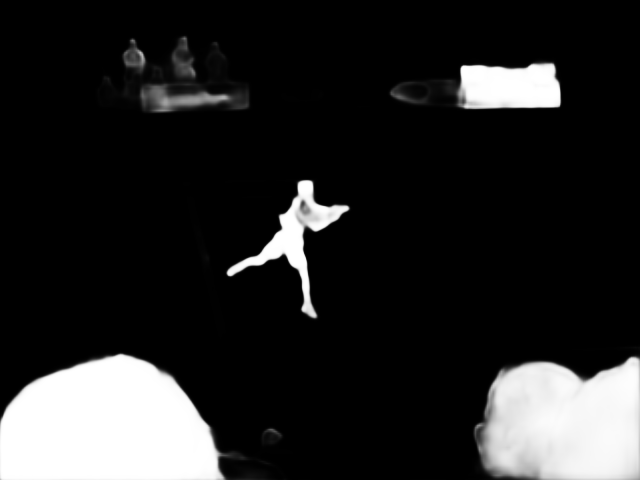}}
\hspace{0.1mm}
\subfloat[ANDA~\cite{ruiz2019anda}]{\includegraphics[width=0.17\linewidth]{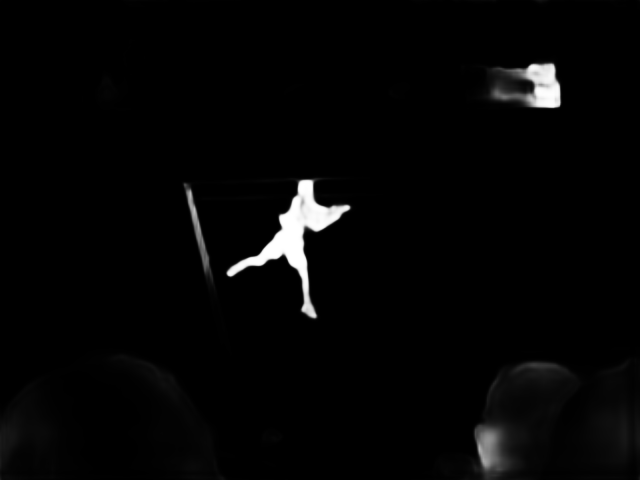}}

\caption{Qualitative difference between segmentation performed with different training data augmentation techniques. An example of each testing dataset is presented. IDA* refers to the best overall model and stands for the data augmentation model that uses the synthetic sample of IDA, our method, Random Horizontal Flip, and the GridMask method. DUT-O* is an abbreviation of DUT-OMRON. DUTS-TE* is an abbreviation of DUTS-TEST. GT is an abbreviation of Ground Truth.}
\label{fig:qualitativeImprovement}       
\end{figure}

An example of each testing dataset and the qualitative difference between segmentation performed with different training data augmentation techniques is presented in Fig.~\ref{fig:qualitativeImprovement}. Three different approaches are illustrated: the best overall model for the data augmentation that uses the synthetic sample of IDA, our method, Random Horizontal Flip, and the GridMask method (Fig.~\ref{fig:qualitativeImprovement} third column); the standard data augmentation technique, horizontal flip (Fig.~\ref{fig:qualitativeImprovement} fourth column); and the one produced by the previous technique ANDA (Fig.~\ref{fig:qualitativeImprovement}  fifth column). 

Note how using the model with IDA, the resulting segmentation improves both in reducing false positives as in the SOC example~(Fig.~\ref{fig:qualitativeImprovement} last row) and increasing true positives like in the HKU-IS~(Fig.~\ref{fig:qualitativeImprovement} fourth row). However, it still produced some false negatives with a low confidence level, represented by a low grayscale value. This is exemplified at the subfigure of THUR15K~(Fig.~\ref{fig:qualitativeImprovement} first row).


\section{Conclusion}\label{sec:end}

In this paper, we propose a new data augmentation technique named IDA in the context of SOD. It uses a linear combination of two different images, the resulting image contains in the foreground a salient object segmented from its original background, affinely transformed, and a full background created using image inpainting to erase its labeled objects. The background choice is based on an inter-image optimization, while object size follows a uniform random distribution within a specified interval, and the object position is intra-image optimal.

During our experiments, two offline (in which images were processed before training started), and two online (in which images were processed per batch during training), data augmentation techniques were analyzed. The offline ones were the ANDA and IDA, while the online ones were the GridMask and random horizontal-flip. The GridMask method has shown an unstable behavior, sometimes improving the results and some times worsening it. Such behavior could be directly tied to its parameterization, in which changing the ratio of the squares delete more or less information. Removing a large chunk of information in a small object can damage the learning process. The methods that achieved the highest results included this policy, meaning that structured information removal can encourage generalization while somewhat unstable. On the other hand, the horizontal flip operation consistently provides an improvement in the results and has a single parameter, the probability of application.

Training a neural network model with synthetic images generated by IDA and other online data augmentation techniques provided a consistent improvement of the S-measure and F-score results in six out of seven test datasets. The achieved results surpassed the previously reported results, setting a new state-of-the-art. These results show that our method is a relevant contribution to the data augmentation training techniques. Furthermore, it can be combined with other online data augmentation techniques providing a more robust policy that can improve deep neural networks training due to their introduced information variance, which encourages the model to learn more generic features and reduce overfitting.


\section*{Acknowledgment}
The authors thank the Coordination for the Improvement of Higher Education Personnel
(CAPES) for the Masters scholarship. We gratefully acknowledge the founders of the publicly
available datasets and the support of NVIDIA Corporation with the donation of the GPUs used for this research.